\documentclass[10pt,twocolumn,letterpaper]{article}

\usepackage[pagenumbers]{iccv} % To force page numbers, e.g. for an arXiv version
\usepackage{multirow}
\usepackage{colortbl}
\usepackage{hhline} 

\definecolor{cvprblue}{rgb}{0.21,0.49,0.74}

\definecolor{seen_back}{RGB}{224, 241, 239}
\definecolor{unseen_back}{RGB}{243, 231, 213}

\usepackage{amsmath}
\usepackage{algorithm}
\usepackage{algorithmic}

\definecolor{iccvblue}{rgb}{0.21,0.49,0.74}
\usepackage[pagebackref,breaklinks,colorlinks,allcolors=iccvblue]{hyperref}

\def\paperID{6846} % *** Enter the Paper ID here
\def\confName{ICCV}
\def\confYear{2025}

\title{Self-Reinforcing Prototype Evolution with Dual-Knowledge Cooperation for Semi-Supervised Lifelong Person Re-Identification}
\author{Kunlun Xu\textsuperscript{\rm 1}~~~ Fan Zhuo\textsuperscript{\rm 1}~~~ Jiangmeng Li\textsuperscript{\rm 2}~~~ Xu Zou\textsuperscript{\rm 3}~~~ Jiahuan Zhou\textsuperscript{\rm 1}\thanks{Corresponding author}  \\
 {\small \textsuperscript{\rm 1} Wangxuan Institute of Computer Technology, Peking University, Beijing, China}\\
 {\small \textsuperscript{\rm 2} University of Chinese Academy of Sciences, Beijing, China}\\
 {\small \textsuperscript{\rm 3} School of Artificial Intelligence and Automation, Huazhong University of Science and Technology, Wuhan, China}\\
 {\small xkl@stu.pku.edu.cn~~~~fanzhuo@stu.xidian.edu.cn~~~~jiangmeng2019@iscas.ac.cn~~~~zoux@hust.edu.cn~~~~jiahuanzhou@pku.edu.cn}\\
}

\begin{document}
\maketitle

\begin{abstract}
Current lifelong person re-identification (LReID) methods predominantly rely on fully labeled data streams. However, in real-world scenarios where annotation resources are limited, a vast amount of unlabeled data coexists with scarce labeled samples, leading to the Semi-Supervised LReID (Semi-LReID) problem where LReID methods suffer severe performance degradation. Existing LReID methods, even when combined with semi-supervised strategies, suffer from limited long-term adaptation performance due to struggling with the noisy knowledge occurring during unlabeled data utilization. In this paper, we pioneer the investigation of Semi-LReID, introducing a novel \textbf{S}elf-Reinforcing \textbf{PR}ototype \textbf{E}volution with \textbf{D}ual-Knowledge Cooperation framework (SPRED). Our key innovation lies in establishing a self-reinforcing cycle between dynamic prototype-guided pseudo-label generation and new-old knowledge collaborative purification to enhance the utilization of unlabeled data. Specifically, learnable identity prototypes are introduced to dynamically capture the identity distributions and generate high-quality pseudo-labels. Then, the dual-knowledge cooperation scheme integrates current model specialization and historical model generalization, refining noisy pseudo-labels. Through this cyclic design, reliable pseudo-labels are progressively mined to improve current-stage learning and ensure positive knowledge propagation over long-term learning. 
Experiments on the established Semi-LReID benchmarks show that our SPRED achieves state-of-the-art performance. Our source code is available at 
\href{https://github.com/zhoujiahuan1991/ICCV2025-SPRED}{https://github.com/zhoujiahuan1991/ICCV2025-SPRED}

% , significantly outperforming existing approaches
% Current lifelong person re-identification (LReID) methods predominantly focus on fully labeled data streams, overlooking the practical challenge of limited annotation resources in realistic applications. This oversight leads to two dilemmas: insufficient new knowledge acquisition from sparsely labeled data and exacerbated catastrophic forgetting due to overfitting. We pioneer the investigation of Semi-Supervised LReID (Semi-LReID), introducing a novel Self-Reinforcing Prototype Evolution with Dual-Knowledge Cooperation framework (SPRED). Our key innovation lies in establishing a self-reinforcing cycle between dynamic prototype-guided pseudo-label generation and cross-model collaborative validation. 

\end{abstract}
\begin{figure}[h!]
		\begin{center}
			\includegraphics[width=1.0\linewidth]{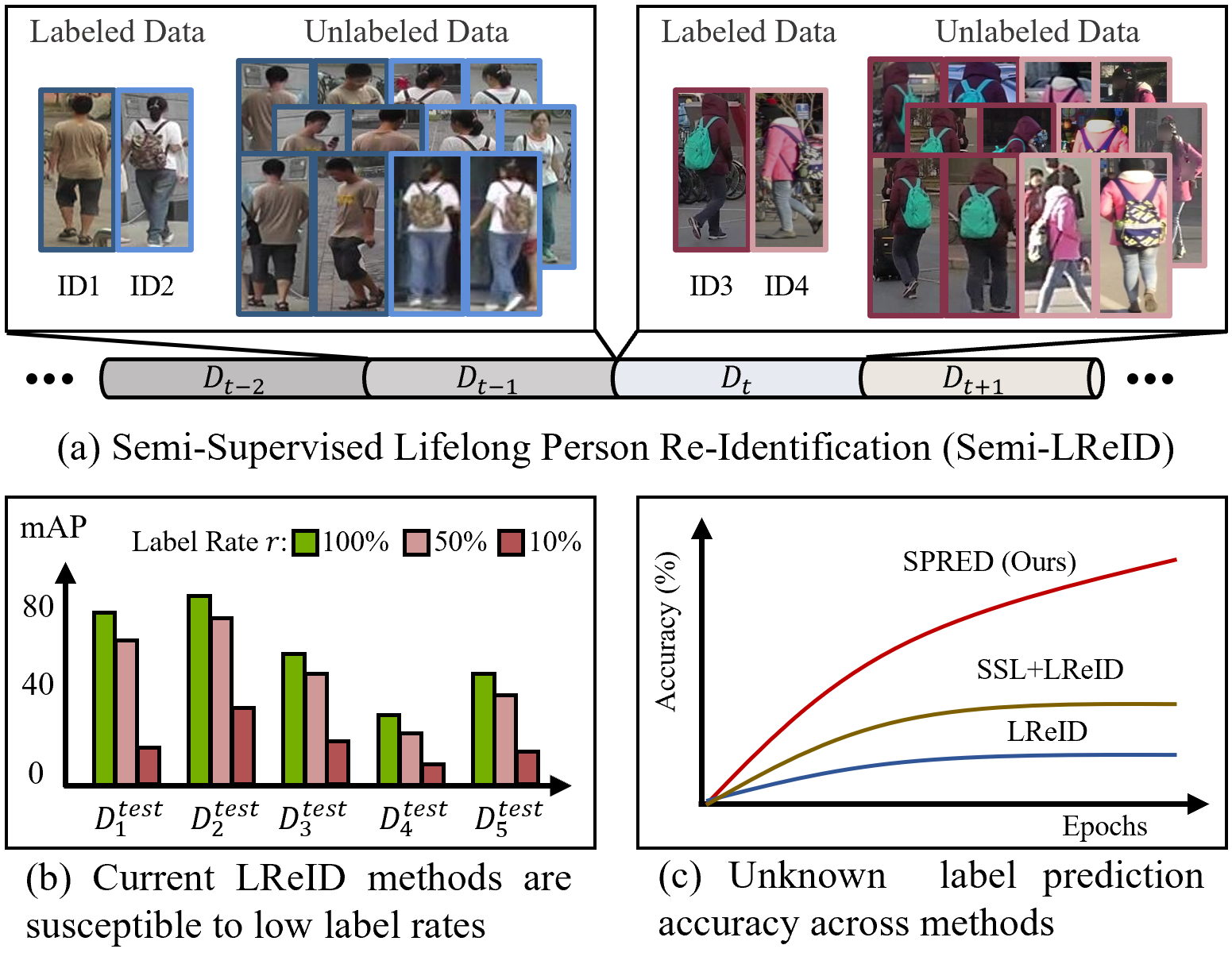}
   % \vspace{-10pt}
			\caption{(a) In Semi-LReID, the training data is continuously obtained with partial labeling. (b) As the labeling rate decreases, the performance of existing  LReID methods suffers severe degradation. (c) Our SPRED improves the knowledge acquisition of unlabeled data significantly.}
   \vspace{-10pt}
			\label{fig:first}
		\end{center}    
	\end{figure}
    
\begin{figure}[h!]
		\begin{center}
			\includegraphics[width=0.98\linewidth]{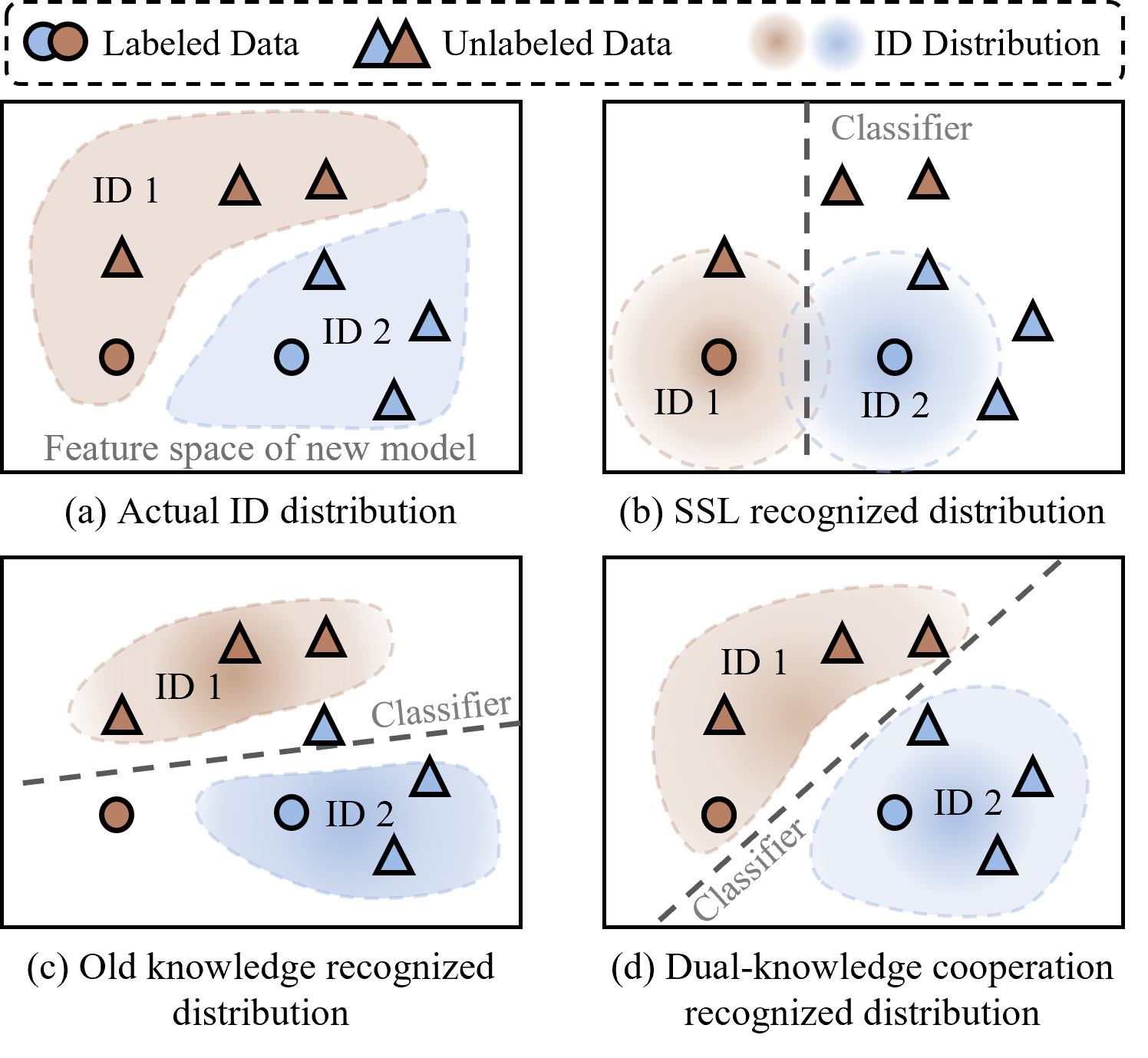}
   % \vspace{-10pt}
			\caption{ (a) Given the labeled and unlabeled data of the new domain, we aim to learn actual ID distributions. (b) SSL tends to approximate ID distributions by emphasizing regions near the labeled data. (c) Old knowledge recognizes samples of the same ID from historical experience (old model). (d) The proposed dual-knowledge cooperation mechanism jointly utilizes the newly learned and complementary historical knowledge to improve ID assignment for unlabeled samples. }
    \vspace{-10pt}
			\label{fig:first-2}
		\end{center}    
	\end{figure}
    \vspace{-5pt}
\section{Introduction}
\label{sec:intro}
Person re-identification (ReID)~\cite{gong2022person,cui2024dma,gongcross, zhong2018camera} has been extensively studied in stationary scenarios~\cite{he2021transreid,fu2022domain,ye2022unsupervised,li2024exemplar}.
Since real-world applications often involve dynamic environments where new data continuously arrives~\cite{wu2021generalising,xu2025componential,zhang2025scap,ge2022lifelong,xu2025long}, lifelong person re-identification (LReID) ~\cite{xu2024distribution,xu2024mitigate,cui2024learning} has emerged as a research frontier, where a significant challenge is the issue of catastrophic forgetting~\cite{xu2024lstkc,cui2025dkc}. Traditional LReID methods typically assume that all training samples are labeled~\cite{pu2021lifelong,xu2024mitigate,yang2023handling}, an assumption that is often impractical due to the high labor costs of ReID annotation~\cite{yin2024robust,shi2024learning,fu2022domain}. To address this limitation, we explore a more realistic setting, \textit{semi-supervised lifelong person re-identification} (Semi-LReID), where only a subset of the data stream is labeled, as depicted in \cref{fig:first} (a). 
In such a scenario, the scarcity of labeled samples leads to significant performance degradation in existing LReID methods, as shown in \cref{fig:first} (b). Consequently, the core challenge in Semi-LReID is to effectively leverage the unlabeled data while mitigating catastrophic forgetting during long-term learning.

A typical approach to tackle Semi-LReID is to adapt existing semi-supervised lifelong learning (SSLL) techniques. However, current SSLL methods~\cite{wang2024persistence, kang2023soft, fan2024dynamic}, usually built on an existing semi-supervised learning baseline~\cite{sohn2020fixmatch}, predominantly develop replay-based anti-forgetting mechanisms. Given the privacy-sensitive nature of human images, storing historical samples for replay is often impractical~\cite{pu2021lifelong, xu2024distribution}. 
Recently proposed replay-free SSLL methods~\cite{gomez2024exemplar} integrate self-supervised pre-training constraints to improve unlabeled data utilization. However, since ReID is an identity-level discrimination task where identity-irrelevant visual cues dominate~\cite{shi2023dual, lin2019bottom}, self-supervised pre-training constraints often exhibit limited robustness~\cite{shi2025multi}.

An alternative approach is to integrate semi-supervised learning (SSL) techniques~\cite{zheng2022simmatch, shi2023dual, wu2023rewarded} into LReID frameworks. While it can enhance unlabeled data exploitation and alleviate catastrophic forgetting, the overall performance remains limited. As shown in Fig.~\ref{fig:first-2} (b), SSL methods tend to approximate identity distributions by overemphasizing regions near labeled samples, resulting in erroneous pseudo-labels~\cite{berthelot2019mixmatch,sohn2020fixmatch}. These noisy labels, in turn, propagate errors during training and hinder the adaptation of the model to new data~\cite{xu2024mitigate}, ultimately diminishing knowledge acquisition capacity in long-term learning.

To address these limitations, we propose a novel Semi-LReID framework, Self-Reinforcing Prototype Evolution with Dual-Knowledge Cooperation (SPRED). As shown in \cref{fig:first-2} (b), our core idea is to jointly leverage newly learned knowledge (Fig.~\ref{fig:first-2} (b)) and complementary historical knowledge (Fig.~\ref{fig:first-2} (c)) to form a Dual-Knowledge Cooperation mechanism (Fig.~\ref{fig:first-2} (d)) that enhances reliable pseudo-label generation. Specifically, SPRED comprises two key components: Dual-Knowledge Cooperation-driven Pseudo-label Purification (DKCP) and Self-Reinforcing Prototype Evolution (SPE). SPE employs learnable identity prototypes to dynamically model identity distributions from pseudo-labels and limited ground-truth annotations. It incorporates a neighbor prototype labeling strategy where ambiguous predictions are discarded and high-quality pseudo-labels are generated. DKCP clusters new data using both the current and historical models and selects reliable pseudo-labels based on clustering consistency among unlabeled samples. These selected pseudo-labels, in turn, motivate the identity prototypes learning in SPE, forming a self-reinforcing cycle that iteratively improves pseudo-label quality and prototype learning. Besides, a prototype structure-based knowledge distillation loss is developed to mitigate catastrophic forgetting, further improving the long-term knowledge consolidation capacity.
Extensive experiments on established Semi-LReID benchmarks demonstrate that SPRED significantly enhances unlabeled data utilization and long-term knowledge accumulation compared to existing solutions.

    In summary, the contributions of this work are three-fold: 
(1) We present a pioneering study on the Semi-LReID task, identifying the limitations of current methods in this setting and introducing a dedicated Semi-LReID benchmark for future research. (2) A novel Semi-LReID method, SPRED, is proposed, forming a cyclically evolved unlabeled data utilization mechanism. The learnable identity prototypes are designed to dynamically enhance pseudo-label generation, while the complementary knowledge from the old model is integrated with new knowledge to purify pseudo-labels. (3) Extensive experiments demonstrate the effectiveness of SPRED in both unlabeled data utilization and long-term knowledge accumulation within the challenging Semi-LReID scenarios.

\begin{figure*}[ht]
    \centering
	\includegraphics[width=1.0\linewidth]{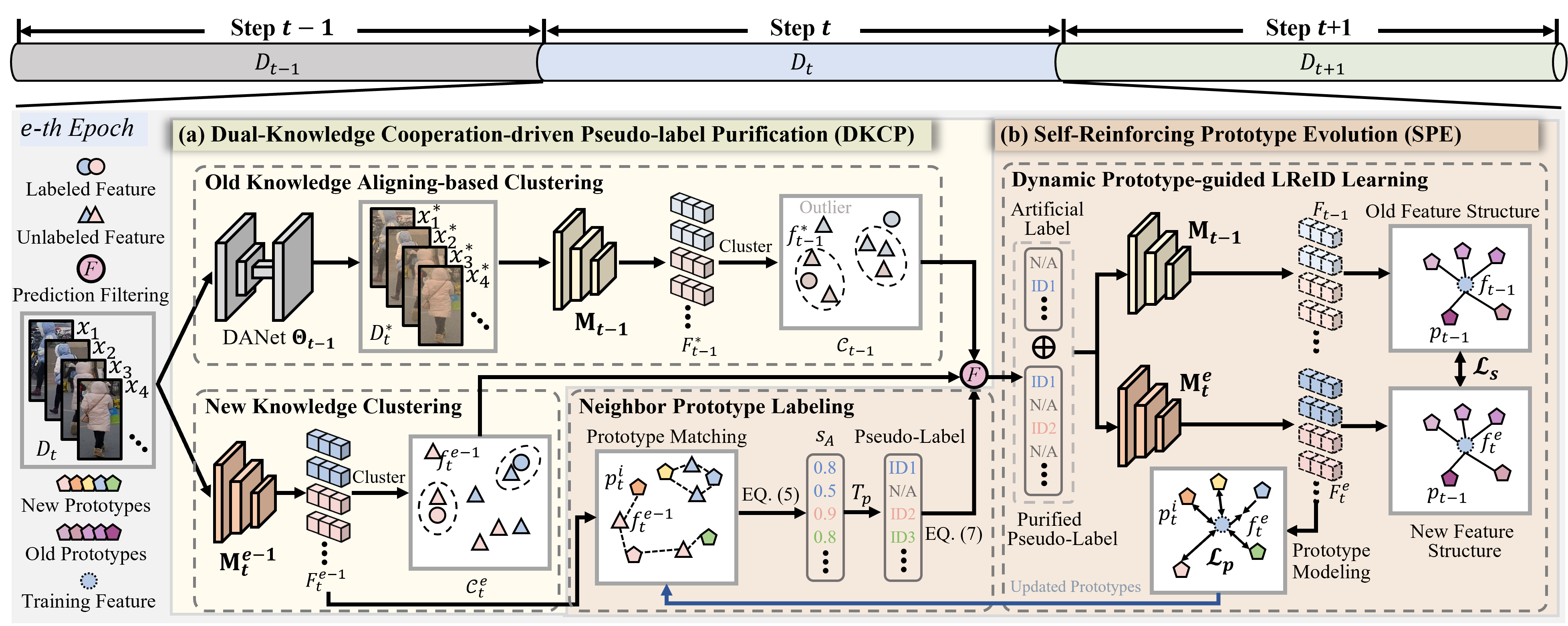}
 
        \caption{\label{fig:framework} The overview of our SPRED. Given a dataset $D_t$, two modules operate cyclically throughout each training epoch $e$. (a) The DKCP mechanism refines the pseudo-labels assigned to unlabeled samples, which are initially generated by (b) the SPE scheme. This scheme introduces learnable prototypes to facilitate LReID training using both artificial labels and pseudo-labels while leveraging the updated prototypes to assign pseudo-labels in the subsequent epoch.
        }        
\end{figure*}
	
\section{Related Work}
\subsection{Lifelong Person Re-Identification}
Lifelong person re-identification~\cite{wu2021generalising,pu2021lifelong,xu2025dask} aims to develop a person retrieval model capable of adapting to multiple domains by learning from non-stationary data. The primary challenge in LReID is the catastrophic forgetting problem~\cite{pu2022meta}, where previously acquired discriminative knowledge is overwritten as new data is introduced~\cite{pu2023memorizing,yu2023lifelong,xu2024distribution,xu2024lstkc}. Existing LReID methods typically assume access to abundant labeled data during training. However, when labeled data is limited, their performance deteriorates significantly, highlighting the need for effective strategies to leverage unlabeled data during lifelong learning.

\subsection{Semi-Supervised Learning}
Semi-supervised learning (SSL) is widely applied in classification and person re-identification tasks where vast images are available, but only a small fraction of the samples are labeled~\cite{shi2023dual}.
Pseudo-label methods aim to predict labels for unannotated data, thereby expanding the labeled training set~\cite{shi2023dual,wu2023rewarded, wei2024semi,chen2023boosting, sosea2023marginmatch}. However, these methods are prone to overfitting artificial labels, leading to noisy pseudo-labels. Such noisy data can introduce erroneous knowledge, which not only degrades the performance of new data but also exacerbates catastrophic forgetting~\cite{xu2024mitigate}.
Self-supervised-based methods employ both labeled and unlabeled data, using supervised and self-supervised supervision, respectively~\cite{fini2023semi, zhai2019s4l}. These approaches typically require large-scale datasets, making them less suitable for lifelong learning scenarios where training data is often limited~\cite{pu2021lifelong}.
Consistency regularization methods enforce consistency in the model's predictions across different perturbations of the same input~\cite{li2021comatch, sohn2020fixmatch, berthelot2019mixmatch, liu2021certainty, sajjadi2016regularization}.
These approaches focus on intra-instance consistency but neglect inter-instance relation learning. However, in Semi-LReID, the inter-instance relations are crucial during testing. As a result, these methods tend to yield sub-optimal results in Semi-LReID scenarios.

\subsection{Semi-Supervised Lifelong Learning}
Semi-supervised Lifelong learning (SSLL) considers a more realistic scenario where only a limited number of samples are annotated in the continual data stream~\cite{fan2024dynamic}. Thus, the new knowledge learning and catastrophic forgetting problems are both challenging.
Existing SSLL works primarily focus on the classification~\cite{wang2021ordisco, luo2024learning, brahma2021hypernetworks, smith2021memory} task. They usually exploit the SSL technique to learn from new data and adopt a replay buffer to store historical exemplars, which are used to mitigate forgetting~\cite{fan2024dynamic, kang2023soft}. However, since human images are privacy-sensitive, these technologies are impractical in real scenarios~\cite {yang2023handling}. 
Recently, a replay-free SSLL method, HDC~\cite{wei2024semi}, was proposed, where the self-supervised pre-training constraints are adopted to
 utilize unlabeled data.  Since ReID is an identity-specific discrimination task, and identity-irrelevant visual cues usually dominate image information.
Self-supervised pre-training constraints often exhibit limited robustness~\cite{shi2023dual, lin2019bottom}.
Therefore, in this paper, we provide a pioneering investigation on the semi-supervised lifelong person re-identification task and propose an exemplar-free solution for realistic applications.

\section{The Proposed Method}

\subsection{Problem Definition}
In Semi-LReID, a set of $T$ datasets $\mathcal{D}=\{D_t\}_{t=1}^{T}$ are given in order. Each dataset $D_t$ contains an unlabeled subset $X_t^u$ and a labeled subset $(X_t^l,Y_t^l)$. The identity labels between different training datasets are disjoint, \textit{i.e.}, $Y_t^l\cap Y_s^l=\emptyset$ when $t\neq s$.
During the $t$-th training step, only $D_t$ is available, the previous and later datasets are inaccessible. Besides, a set of $T$ test datasets $\mathcal{D}^{te}=\{D_t^{te}\}_{t=1}^{T}$ is collected to evaluate the long-term knowledge accumulation capacity of the model. In addition, $U$ datasets $\mathcal{D}^{un}=\{ D^{un}_i\}_{i=1}^{U}$ from novel domains are exploited to evaluate the generalizability of the model.

\subsection{Overview}
As shown in Fig.~\ref{fig:framework}, the proposed SPRED method comprises two main components: a Dual-Knowledge Cooperation-driven Pseudo-label Purification (DKCP) mechanism and a Self-Reinforcing Prototype Evolution (SPE) scheme. SPE contains a Dynamic Prototype-guided LReID Learning module (DPL) and a Neighbor Prototype Labeling module (NPL), which aim to conduct dynamic prototype modeling and prototype-based pseudo-label generation, respectively.  
DKCP consists of an Old Knowledge Aligning-based Clustering module (OKAC) and a New Knowledge Clustering module (NKC), which adopt the new and old models to generate new sample clusters, respectively. The clustering results are exploited to filter unreliable pseudo-labels. 
% In the following, we introduce the above modules in the order of DPL, NPL, and DKCP.
\subsection{Self-Reinforcing Prototype Evolution}
In this section, we first present the learnable prototype modeling process in the Semi-LReID scenario, followed by a detailed explanation of the prototype-based pseudo-label generation strategy.

\textbf{Dynamic Prototype-guided LReID Learning}: At training step $t$, given $L_t$ identities, a set of learnable prototypes $\mathcal{P}_t=\{p_t^i\}_{i=1}^{L_t}$ is adopted, where each $p_t^i \in \mathbb{R}^d$ is a $d$-dimensional vector corresponding to a specific identity.
Additionally, we exploit a set of valid pseudo labels, denoted $Y_t^{pse}$, with an associated image set $X_t^{pse}$. At the $e$-th epoch, a new LReID model $\boldsymbol{\mathrm{M}}_t^e$ is trained on these data. The final LReID model learned from $D_t$ is denoted as $\boldsymbol{\mathrm{M}}_t$.

New Knowledge Learning:
Given an input image $x$ which is sampled from $X_t^l \cup X_t^{pse}$, $\boldsymbol{\mathrm{M}}_t^e$ is adopted to extract the feature $f_t^e$. Then a prototype-orient identity learning loss is introduced, which is calculated as follows:
\begin{equation}
   \mathcal{L}_{p}=-\log \frac{e^{<f_t^e, p_t^x>}}{\sum_{i=1}^{L_t} e^{<f_t^e, p_t^i>}}
    \label{eq:proto-loss},
\end{equation}
where $p_t^x$ denotes the prototype sharing the same label as $x$. As illustrated in \cref{fig:framework} (b), \cref{eq:proto-loss} encourages the instance feature to align more closely with its corresponding identity prototype while distancing itself from other prototypes. Since both $\mathcal{P}_t$ and $\boldsymbol{\mathrm{M}}_t^e$ are learnable, the prototypes and model parameters are updated dynamically during training.

To further enhance feature learning in LReID, we also integrate the widely-used Triplet loss $\mathcal{L}_{Tri}$~\cite{xu2024lstkc}. The total loss for new knowledge learning is thus formulated as:
\begin{equation}
   \mathcal{L}_{base}=\mathcal{L}_{p}+\mathcal{L}_{Tri}
    \label{eq:base_loss}.
\end{equation}

Old Knowledge Non-Forgetting:
Since $\mathcal{L}_{p}$ guides the model to learn one-instance \vs all-prototype similarity structures, we aim to preserve the structural knowledge learned by the old model to mitigate catastrophic forgetting. Specifically, let $f_{t-1} \in \mathbb{R}^d$ denote the feature extracted by $\boldsymbol{M}_{t-1}$. The  prototype-based old structure is represented as:
\begin{equation}
S(f_{t-1}, \mathcal{P}_{t-1})=[g(f_{t-1}, \mathcal{P}_{t-1}, 1),...,g(f_{t-1}, \mathcal{P}_{t-1}, N_{t-1})]
\end{equation}
where $g(f_{t-1}, \mathcal{P}_{t-1}, j)=\frac{e^{<f_{t-1}, p_{t-1}^j>}}{\sum_{i=1}^{L_{t-1}} e^{<f_{t-1}, p_{t-1}^i>}}$ represents the affinity between input image $x$ and each prototype $p_{t-1}^j$ according to the old knowledge. 

Similarly, the prototype-based new structure  $S(f_{t}^e, \mathcal{P}_{t-1})$ could be obtained. Then, to alleviate structural drift as the new model evolves, we introduce a prototype-guided feature structure maintaining loss:
\begin{equation}
   \mathcal{L}_{s}=\mathcal{L}_{KL}\big(S(f_{t-1}, \mathcal{P}_{t-1})||S(f_{t}^e, \mathcal{P}_{t-1})\big)
    \label{eq:structure-loss},
\end{equation}
where $\mathcal{L}_{KL}$ is the Kullback Leibler divergence~\cite{hershey2007approximating}.

\textbf{Neighbor Prototype Labeling}:\label{sec:NPL}
Existing pseudo-label generation methods generally rely on classification scores and a threshold to create pseudo-labels~\cite{fini2023semi,chen2023learn}, where the correct prediction score is susceptible to multiple neighbor classes, causing many informative samples to be discarded.

To address this issue, we propose to focus on the top-2 nearest classes to obtain the pseudo-labels and ignore the less-related classes, as shown in \cref{fig:framework} (b). Specifically, given the image feature $f_t^{e-1}\in\mathbb{R}^d$ extracted by $\boldsymbol{\mathrm{M}}_t^{e-1}$, we obtain the top-2 nearest prototypes, where the nearest and the second nearest prototypes are represented as $p_A$ and $p_B$, respectively. The classification score of the $p_A$ class is obtained by:
\begin{equation}
    s_A=\frac{e^{<f_t^{e-1}, p_A>}}{e^{<f_t^{e-1}, p_A>}+e^{<f_t^{e-1}, p_B>}}
    \label{eq:neighbor-score}.
\end{equation}
Then, we select the unlabeled samples with a threshold $T_p$. Therefore, the predicted label $l_i$ of a sample $x_i$ can be obtained by:
\begin{equation}
    l_i=\left\{
		\begin{aligned}
			&  y_i \qquad x_i \in X_t^l\\
			& l_A  \qquad x_i \notin X_t^l, s_A>T_p\\
            &N/A   \quad x_i \notin X_t^l, s_A\leq T_p   
		\end{aligned}
		\right.,
    \label{eq:neighbor-score}
\end{equation}
where $y_i$ denotes the artificial label of $x_i$, $l_A$ represents the corresponding label of $p_A$, and $N/A$ means the pseudo-label is unavailable according to the prototypes. 
\subsection{Dual-Knowledge Cooperation-driven Pseudo-label Purification}
The pseudo-labels generated in \cref{sec:NPL}  are inevitably noisy due to the prototypes overfitting the labeled data, which can misguide model training and exacerbate catastrophic forgetting~\cite{xu2024mitigate}. To settle this, we propose to leverage both new and old knowledge from the current and previous models to filter noisy pseudo-labels, thereby enhancing the effective utilization of unlabeled samples.

\textbf{ Old Knowledge Aligning-based Clustering}:
Due to the domain shift across training steps, directly applying $\boldsymbol{\mathrm{M}_{t-1}}$ to process $D_t$ leads to suboptimal results. To settle this, we employ a distribution alignment network (DANet) $\boldsymbol{\Theta}_{t-1}$ to map the new data $D_t$ to the distribution of $D_{t-1}$. The transformed data are denoted as $D_t^*=\{x_i^*\}_{i=1}^{n_t}$, where $n_t$ is the number of images in $D_t$. The architecture and training protocol for DANet is based on~\cite{gu2023color}, with additional details available in our Supplementary Material.

Then, the old model $\boldsymbol{\mathrm{M}_{t-1}}$ is exploited to extract image features $F_{t-1}^* = \{{f_{t-1}^{(i)}}^*\}_{i=1}^{|D_t^*|}$ from the transformed data $D_t^*$. Subsequently, the DBSCAN clustering algorithm~\cite{gu2023color} is applied to $F_{t-1}^*$, resulting in $N_C^{t-1}$ clusters, denoted as $\mathcal{C}_{t-1} = \{C_1^{t-1}, C_2^{t-1}, \dots, C_{N_C^{t-1}}^{t-1}\}$. Each cluster $C_i^{t-1}$ represents a group of instances exhibiting similar patterns based on the knowledge encoded in $\boldsymbol{\mathrm{M}_{t-1}}$.

\textbf{New Knowledge Clustering}: 
Given the new data $D_t$, the new model $\boldsymbol{\mathrm{M}_{t}^{e-1}}$ is utilized to extract features $F_t^{e-1}=\{f_{t}^{(i),e-1}\}_{i=1}^{N_t}$.
Next, the clustering algorithm is also applied to process $F_t^{e-1}$ and a set of $N_C^{t,e-1}$ clusters 
$\mathcal{C}_{t}^{e-1}=\{C_1^{e-1},C_2^{e-1},...,C_{N_C^{t,e-1}}^{e-1}\}$ is obtained. Here, each cluster $C_i^{e-1}$ is composed of instances whose features extracted by the new model are highly similar to each other.
% Note the new knowledge clustering and 

\textbf{Noisy Pseudo-Label Filtering}: 
As shown in \cref{fig:framework} (a), we aim to utilize the clustering results generated by new and old models to filter the noisy pseudo-labels.
Firstly, to bridge the clustering results and pseudo-labels, we split the new samples into sets according to the pseudo-labels generated in \cref{sec:NPL}, 
resulting in a collection $\mathcal{S}^{e-1}=\{S_1^{e-1},S_2^{e-1},...,S_{L_t}^{e-1}\}$, where each subset $S_i^{e-1}=\{x_j: x_j\in X_t^l \cup X_t^u, l_j=i\}$ is set of samples with the same pseudo-label.

For an unlabeled instance $x$, pseudo-label set $\mathcal{S}$ and clustering result $\mathcal{C}$, we define the set-based label confidence score $LC_s(x, \mathcal{S}, \mathcal{C})$ which is calculated by 
\begin{equation}
    LC_s(x,\mathcal{S}, \mathcal{C})=\frac{S_x\cap C_x}{S_x\cup C_x}
    \label{eq:consistency-score},
\end{equation}
where $S_x$ and $C_x$ are the elements of $\mathcal{S}$ and $\mathcal{C}$ respectively, which both contain $x$. 

Given the new knowledge clustering result $\mathcal{C}_{t}^{e-1}$ and an unlabelled sample $x$,
a threshold $T_c$ is adopted to obtain the purified pseudo-label set $Y_{e-1}^{pse}=\{y^{pse}_{i,e-1}: x_i\in X_t^u\}$ where $y^{pse}_i$ is obtained by
\begin{equation}
   y^{pse}_{i,e-1}=\left\{
		\begin{aligned}
			& l_i \qquad\ \ LC_s(x_i,\mathcal{S}^{e-1}, \mathcal{C}_t^{e-1})>T_c\\
			& N/A  \quad LC_s(x_i,\mathcal{S}^{e-1}, \mathcal{C}_t^{e-1})\leq T_c
		\end{aligned}
		\right.
        \label{eq:pseudo_e-1}
\end{equation}

Similarly, given the old knowledge clustering result $\mathcal{C}_{t-1}$, we can obtain another purified pseudo-label set $Y_{t-1}^{pse}=\{y^{pse}_{i,t-1}: x_i\in X_t^u\}$ by introducing a threshold $T_o$, where $y^{pse}_{i,t-1}$ is obtained following \cref{eq:pseudo_e-1}.

Compared to using a purified pseudo-label set alone or using their intersection, we experimentally found that using $Y_{e-1}^{pse}$ and $Y_{t-1}^{pse}$ results in the highest performance, \textit{i.e.}, the pseudo label for LReID model training is obtained by
\begin{equation}
    Y_t^{pse}=Y_{t-1}^{pse} \cup Y_{e-1}^{pse}
    \label{eq:pseudo-label}.
\end{equation}
This underscores that the new and old model clustering results are complementary to each other.

\subsection{Training and Inference}
When training on the first dataset, only $\mathcal{L}_{base}$ is exploited as the supervision.
When training on the subsequent datasets, the overall loss is calculated by:
 \begin{equation}
   \mathcal{L}=\mathcal{L}_{base} +\alpha \mathcal{L}_{s}
    \label{eq:overall_loss},
\end{equation}
where $\alpha$ is a hyperparameter to balance the new knowledge learning and old knowledge forgetting. 

During inference, following the previous works~\cite{xu2024distribution, sun2022patch}, the image features extracted by the final model $\boldsymbol{\mathrm{M}}_{T}$ are adopted for person retrieval. 

	\begin{table*}[h]
		\centering
		      
              \setlength{\tabcolsep}{1.0mm}{
                \begin{tabular}{cclcccccccccc>{\columncolor{seen_back}}c>{\columncolor{seen_back}}c>{\columncolor{unseen_back}}c>{\columncolor{unseen_back}}cccc}   
				\hline
                % \multirow{2}[1]{*}{\rotatebox{90}{\makecell{Label\\ Rate}}} 
                \multirow{2}[1]{*}{$r$} 
                % & \multirow{2}[1]{*}{\rotatebox{90}{Type}} 
                & \multirow{2}[1]{*}{Type} 
                & \multirow{2}[1]{*}{Method} & 
                \multicolumn{2}{c}{Market-1501} & \multicolumn{2}{c}{CUHK-SYSU} & 
                \multicolumn{2}{c}{LPW} & \multicolumn{2}{c}{MSMT17} & 
                \multicolumn{2}{c}{CUHK03} & \multicolumn{2}{>{\columncolor{seen_back}}c}{\textbf{Seen-Avg}} & 
                \multicolumn{2}{>{\columncolor{unseen_back}}c}{\textbf{UnSeen-Avg}} \\

                \hhline{~~~----------------}
				% \cline{4-17} 
                & &   & mAP & R@1  & mAP  & R@1  & mAP  & R@1  & mAP  & R@1  & mAP  & R@1  & mAP  & R@1  & mAP  & R@1 \\    
				\hline
				\multirow{11}[1]{*}{\rotatebox{90}{50\%}} &\multirow{3.3}[0]{*}{\rotatebox{90}{LReID}} &
                PatchKD~\cite{sun2022patch}  &65.8	&84.5	&73.0	&75.9	&25.7	&34.4	&5.3	&14.7	&19.3	&19.7	&37.8	&45.8	&40.3	&34.1
                \\		
                % & & CKP~\cite{xu2024lstkc}  &55.2  &78.5  &81.0  &82.8  &44.3  &56.9  &17.6  &40.5  &39.2  &40.9  &47.5  &59.9  &55.7  &48.1  \\
				& & DKP~\cite{xu2024distribution}     &49.5	&73.9	&80.7	&83.2	&35.4	&47.6	&14.4	&33.9	&26.6	&26.3	&41.3	&53.0	&52.8	&46.6

                \\
				& & LSTKC~\cite{xu2024lstkc} &48.4	&71.4	&80.8	&83.4	&39.3	&52.0	&16.7	&38.2	&37.0	&38.1	&44.4	&56.6	&53.3	&45.7

  \\
				% \cline{2-17}
                \hhline{~----------------}
				&\multirow{5.3}[0]{*}{\rotatebox{90}{SSL+LReID}} & 
    
				CDMAD${}^\dag$~\cite{lee2024cdmad} &42.0	&70.5	&77.6	&80.7	&21.9	&35.4	&7.4	&23.4	&14.5	&15.1	&32.7	&45.0	&47.4	&40.3

\\
				&& ShrinkM${}^\dag$\cite{yang2023shrinking}    &51.9	&75.9	&77.5	&80.6	&42.6	&57.5	&21.7	&49.3	&37.7	&39.4	&46.3	&60.5	&57.9	&\textcolor{blue}{\textbf{50.6}}

 \\
                &&SimMV2${}^\dag$~\cite{zheng2023simmatchv2}  &34.6	&59.3	&63.2	&65.9	&28.0	&41.1	&11.5	&30.3	&29.6	&31.4	&33.4	&45.6	&46.9	&40.3

 \\
                && DPIS${}^\dag$\cite{shi2023dual}  &49.9	&72.1	&82.0	&84.4	&40.8	&53.7	&17.2	&38.8	&35.5	&35.6	&45.1	&56.9	&55.3	&48.5

 \\
                && HDC${}^\dag$\cite{wei2024semi} &51.8	&75.2	&79.7	&82.2	&43.5	&57.0	&22.2	&49.1	&40.1	&42.3	&47.5	&\textcolor{blue}{\textbf{61.2}}	&\textcolor{blue}{\textbf{58.2}}	&\textcolor{blue}{\textbf{50.6}}

\\
				% \cline{3-17}
                \hhline{~———----------------}
                &\multirow{2.3}[0]{*}{\rotatebox{90}{Ours}}
				& \textbf{SPRED} &59.8	&79.7	&82.8	&84.6	&49.7	&60.2	&17.4	&38.2	&40.6	&41.3	&\textcolor{blue}{\textbf{50.1}}	&60.8	&57.8	&50.3
  \\
                &&\textbf{SPRED${}^\ddag$} &59.4	&80.7	&79.6	&81.5	&47.3	&60.2	&21.6	&47.3	&48.4	&49.4	&\textcolor{red}{\textbf{51.3}}	&\textcolor{red}{\textbf{63.8}}	&\textcolor{red}{\textbf{60.7}}	&\textcolor{red}{\textbf{53.9}}

\\

				\hline
				\multirow{11}[1]{*}{\rotatebox{90}{20\%}} &\multirow{3.3}[0]{*}{\rotatebox{90}{LReID}} &
                PatchKD~\cite{sun2022patch} &43.6	&65.7	&67.1	&70.5	&10.0	&15.8	&3.0	&8.8	&7.0	&6.2	&26.1	&33.4	&30.4	&24.8

 \\		
                % & & CKP~\cite{xu2024lstkc} &48.1  &72.7  &78.8  &81.4  &39.2  &53.4  &16.2  &38.5  &29.7  &29.4  &42.4  &55.1  &52.2  &43.8  \\
				& & DKP~\cite{xu2024distribution}  &30.4	&55.8	&73.5	&76.1	&21.5	&31.0	&8.8	&24.1	&8.3	&7.0	&28.5	&38.8	&40.8	&34.8

  \\

				& & LSTKC~\cite{xu2024lstkc}  &34.8	&58.6	&77.1	&80.0	&28.2	&40.4	&12.2	&29.7	&19.4	&18.2	&34.3	&45.4	&47.0	&40.1

  \\

				% \cline{2-17}
                \hhline{~----------------}
				&\multirow{5.3}[0]{*}{\rotatebox{90}{SSL+LReID}} & 
				CDMAD${}^\dag$~\cite{lee2024cdmad} &29.8	&56.9	&74.0	&77.5	&19.1	&31.1	&6.0	&19.9	&9.5	&8.6	&27.7	&38.8	&43.6	&37.0

  \\
				&& ShrinkM${}^\dag$\cite{yang2023shrinking}  &41.7	&68.0	&73.3	&76.4	&34.8	&49.2	&17.5	&42.9	&22.8	&22.4	&38.0	&51.8	&53.0	&45.8

 \\
                &&SimMV2${}^\dag$~\cite{zheng2023simmatchv2}  &34.5	&59.1	&65.1	&68.4	&26.1	&38.1	&10.5	&28.6	&19.0	&19.2	&31.0	&42.7	&45.7	&38.3

  \\
                && DPIS${}^\dag$\cite{shi2023dual} &34.8	&59.3	&77.9	&81.1	&29.3	&41.1	&12.5	&30.8	&18.5	&16.8	&34.6	&45.8	&47.7	&40.7

 \\
                && HDC${}^\dag$\cite{wei2024semi} &39.9	&65.5	&75.3	&78.5	&31.8	&44.8	&16.4	&40.3	&25.3	&24.1	&37.7	&50.6	&52.5	&45.7

 \\
				% \cline{3-17}
                \hhline{~———----------------}
                &\multirow{2.3}[0]{*}{\rotatebox{90}{Ours}}
				& \textbf{SPRED}  &54.7	&75.7	&81.1	&83.0	&45.0	&56.9	&16.1	&36.3	&35.8	&35.4	&\textcolor{blue}{\textbf{46.5}}	&\textcolor{blue}{\textbf{57.5}}	&\textcolor{blue}{\textbf{55.7}}	&\textcolor{blue}{\textbf{48.4}}

   \\
                &&\textbf{SPRED${}^\ddag$} &54.6	&77.1	&78.5	&80.5	&44.9	&58.9	&21.3	&47.5	&38.6	&38.2	&\textcolor{red}{\textbf{47.6}}	&\textcolor{red}{\textbf{60.4}}	&\textcolor{red}{\textbf{58.9}}	&\textcolor{red}{\textbf{51.5}}

  \\
                \hline
				\multirow{11}[1]{*}{\rotatebox{90}{10\%}} &\multirow{3.3}[0]{*}{\rotatebox{90}{LReID}} &
                PatchKD~\cite{sun2022patch} &20.6	&39.0	&58.5	&62.4	&3.5	&6.1	&1.1	&3.7	&3.0	&2.3	&17.3	&22.7	&20.8	&16.5

 \\		
                % & & CKP~\cite{xu2024lstkc}  &40.4  &67.0  &75.3  &78.3  &33.2  &46.2  &13.5  &34.5  &24.3  &24.2  &37.3  &50.0  &47.9  &40.8  \\
				& & DKP~\cite{xu2024distribution}   &17.9	&39.5	&67.5	&70.3	&12.5	&19.0	&5.7	&17.4	&4.9	&3.2	&21.7	&29.9	&32.9	&27.1

   \\
				& & LSTKC~\cite{xu2024lstkc}  &24.0	&46.7	&72.8	&76.8	&20.0	&29.6	&7.6	&20.4	&14.2	&12.1	&27.7	&37.1	&39.5	&32.7

 \\
				% \cline{2-17}
                \hhline{~----------------}
				&\multirow{5.3}[0]{*}{\rotatebox{90}{SSL+LReID}} & 

				CDMAD${}^\dag$~\cite{lee2024cdmad} &23.0	&49.8	&72.2	&75.9	&16.6	&26.4	&5.8	&19.3	&7.8	&6.9	&25.1	&35.7	&42.7	&37.1
 \\
				&& ShrinkM${}^\dag$\cite{yang2023shrinking}  &32.2	&58.1	&70.2	&74.0	&27.7	&42.1	&13.0	&35.1	&18.5	&17.9	&32.3	&45.4	&48.0	&41.2

  \\
                &&SimMV2${}^\dag$~\cite{zheng2023simmatchv2}  &29.2	&54.1	&64.3	&68.0	&20.5	&32.0	&8.5	&24.4	&14.9	&14.1	&27.5	&38.5	&42.3	&35.2

   \\
                && DPIS${}^\dag$\cite{shi2023dual} &25.4	&48.1	&74.0	&77.4	&20.9	&30.6	&8.6	&22.3	&14.0	&12.2	&28.6	&38.1	&41.4	&34.2

 \\
                && HDC${}^\dag$\cite{wei2024semi} &26.0	&48.8	&70.2	&73.9	&20.8	&29.7	&8.1	&23.1	&12.4	&10.9	&27.5	&37.3	&45.6	&39.2

 \\
				% \cline{3-17}	
                \hhline{~———----------------}
                &\multirow{2.3}[0]{*}{\rotatebox{90}{Ours}}
                  & \textbf{SPRED}  &45.9	&69.4	&79.3	&81.6	&40.0	&51.1	&14.2	&33.0	&36.7	&37.1	&\textcolor{blue}{\textbf{43.2}}	&\textcolor{blue}{\textbf{54.4}}	&\textcolor{blue}{\textbf{51.7}}	&\textcolor{blue}{\textbf{44.6}}

 \\
                 & & \textbf{SPRED${}^\ddag$}  &49.5	&74.0	&76.5	&79.1	&40.9	&55.4	&19.4	&45.3	&37.5	&38.4	&\textcolor{red}{\textbf{44.8}}	&\textcolor{red}{\textbf{58.4}}	&\textcolor{red}{\textbf{55.9}}	&\textcolor{red}{\textbf{48.7}}

  \\
				\hline
			\end{tabular}%
		}
        \vspace{-5pt}
        \caption{Results comparison under different label rate $r$. ${}^\dag$ indicates the state-of-the-art SSL method is integrated with the anti-forgetting designs of LSTKC. ${}^\ddag$ represents adopting the data augmentation strategy of ShrinkM.}	        
		\label{tab:setting1}%
	\end{table*}%

\section{Experiments}

\subsection{Datasets and Evaluation Metrics}
In this section, we present a new and large-scale Semi-LReID benchmark for comprehensive model evaluation:
	
\textbf{Datasets}: 12 ReID datasets are adopted to compose our Semi-LReID benchmark, 5 of which are set as training domains (Market1501~\cite{zheng2015scalable}, LPW~\cite{song2018region}, CUHK-SYSU~\cite{xiao2016end}, MSMT17-V2~\cite{wei2018person}, and CUHK03~\cite{li2014deepreid}), and the other 7 smaller datasets are set as test-only domains (CUHK01~\cite{li2012human}, CUHK02~\cite{li2013locally}, VIPeR~\cite{gray2008viewpoint}, PRID~\cite{hirzer2011person}, i-LIDS~\cite{branch2006imagery}, GRID~\cite{loy2010time}, and SenseReID~\cite{zhao2017spindle}). More details of the datasets are provided in our Supplementary Materials.	

\textbf{Label Rate}: Following the previous semi-supervised learning settings~\cite{cai2021exponential,shi2023dual,zheng2022simmatch}, diverse labeling rates (10\%, 20\%, 50\%) are adopted to evaluate the effectiveness of the methods under different scenarios, where the labeled data are sampled randomly. 

\textbf{Evaluation Metrics}:
Following the standard LReID protocols~\cite{pu2021lifelong, xu2024lstkc}, mean Average Precision (mAP) and Rank-1 (R@1) accuracy are adopted to evaluate model performance on each ReID dataset. Moreover, the average mAP and R@1 across datasets are calculated to assess the overall performance on diverse domains.

\subsection{Implementation Details}
Following the previous works~\cite{sun2022patch,xu2024lstkc}, the ResNet-50 is set as the LReID backbone. 
The model is initially trained on the first dataset for 80 epochs, with subsequent datasets trained for 60 epochs each. For optimization, an SGD optimizer with an initial learning rate of $8 \times 10^{-3}$ is used.
% Input images are resized to $256 \times 128$, and the batch size is set to 128.
% The hyperparameters $\alpha$, $T_c$, $T_o$, and $T_p$ are set to 4.0, 0.1, 0.6 and 0.7, respectively. 
To ensure fair comparisons with existing LReID methods, standard LReID data augmentations~\cite{xu2024distribution}, including random cropping, erasing, and horizontal flipping, are applied to obtain the SPRED model. Additionally, we observed that the strong-weak augmentation strategy commonly used in SSL~\cite{yang2023shrinking}, can boost generalization in Semi-LReID. Therefore, we incorporate this SSL augmentation strategy, with further details provided in the supplementary material, to obtain SPRED${}^\ddag$. 
The experiments are conducted on the label rate $r$=10\% by default.
% All experiments are conducted on a single NVIDIA 4090 GPU.

% To ensure fair comparisons with existing LReID methods, we apply standard LReID data augmentations~\cite{xu2024distribution}, including random cropping, erasing, and horizontal flipping to obtain our SPRED model. Additionally, we observed that the strong-weak augmentation strategy, widely used in SSL~\cite{yang2023shrinking}, enhances generalization on the Semi-LReID benchmark. Thus, we incorporate this SSL augmentation strategy which is detailed in the supplementary. The obtained model is noted as SPRED$\ddag$.
% All experiments are conducted on a single NVIDIA 4090 GPU.  
% Following~\cite{xu2024mitigate}, the label update is conducted every 5 epochs for the sake of effectiveness and efficiency.
\subsection{Compared Methods}
We compare the proposed method with two categories of approaches: A. \textit{LReID} methods, including PatchKD~\cite{sun2022patch}, LSTKC~\cite{xu2024lstkc}, and DKP~\cite{xu2024distribution}. B. \textit{SSL} methods, including SSL-Classification methods, CDMAD~\cite{lee2024cdmad}, ShrinkM~\cite{yang2023shrinking}, SimMV2~\cite{zheng2023simmatchv2} and SSL-ReID methods, DPIS~\cite{shi2023dual}, HDC~\cite{wei2024semi}. Since SSL methods do not incorporate an anti-forgetting mechanism, the anti-forgetting strategy of the state-of-the-art LReID method (LSTKC~\cite{xu2024lstkc}) is integrated into these SSL approaches. The resulting methods are referred to as \textit{SSL+LReID} approaches. All experimental results are obtained using the official implementation or following the official paper, with a parameter grid search conducted to ensure optimal performance.

The performance of different methods on each seen domain is shown in \cref{tab:setting1}. Besides, we also report the average performance across all seen domains (Seen-Avg) and unseen domains (UnSeen-Avg). The best and second best results are highlighted in \textcolor{red}{\textbf{Red}} and \textcolor{blue}{\textbf{Blue}}, separately.

\subsection{Seen-Domain Performance Evaluation}
\textbf{Compared to LReID Methods}: 
In \cref{tab:setting1}, our SPRED consistently outperforms existing LReID methods in terms of Seen-Avg mAP/R@1 across different labeling rates. Specifically, compared to the state-of-the-art LSTKC, SPRED achieves notable improvements of \textbf{5.7\%/4.2\%}, \textbf{12.2\%/12.1\%} and \textbf{15.5\%/17.3\%} in Seen-Avg mAP/R@1 as the labeling rate decreases from 50\% to 10\%. These results verify the effectiveness of SPRED in label-limited settings, primarily due to its dynamic prototype learning and dual-knowledge cooperation mechanisms, which enhance the utilization of unlabeled data.

\textbf{Compared to SSL+LReID Methods}: 
As shown in \cref{tab:setting1}, our SPRED surpasses existing SSL+LReID approaches, achieving at least \textbf{8.5\%/5.7\%} and \textbf{10.9\%/9.0\%} improvements in Seen-Avg mAP/R@1 under 20\% and 10\% labeling rates, respectively. Under a 50\% labeling rate, SPRED demonstrates a 2.6\% increase in Seen-Avg mAP but a 0.4\% decrease in Seen-Avg R@1 compared to HDC. This higher R@1 performance by HDC primarily stems from its data augmentation strategy, which enhances input data diversity to facilitate LReID training.

By incorporating the widely used data augmentation strategy of SSL~\cite{yang2023shrinking}, our enhanced SPRED${}^\ddag$ model surpasses all SSL+LReID methods in Seen-Avg mAP/R@1 metrics across different labeling rates. Specifically, we observe improvements of \textbf{3.8\%/2.6\%}, \textbf{9.6\%/8.6\%} and \textbf{12.5\%/13.0\%} in Seen-Avg mAP/R@1 at 50\%, 20\%, and 10\% label rates, respectively. This superior performance is attributed to our self-reinforcing prototype evolution framework, which enhances pseudo-label quality and boosts the capacity of the model for new knowledge acquisition.

\begin{figure}[t]
    \centering
	\includegraphics[width=1\linewidth]{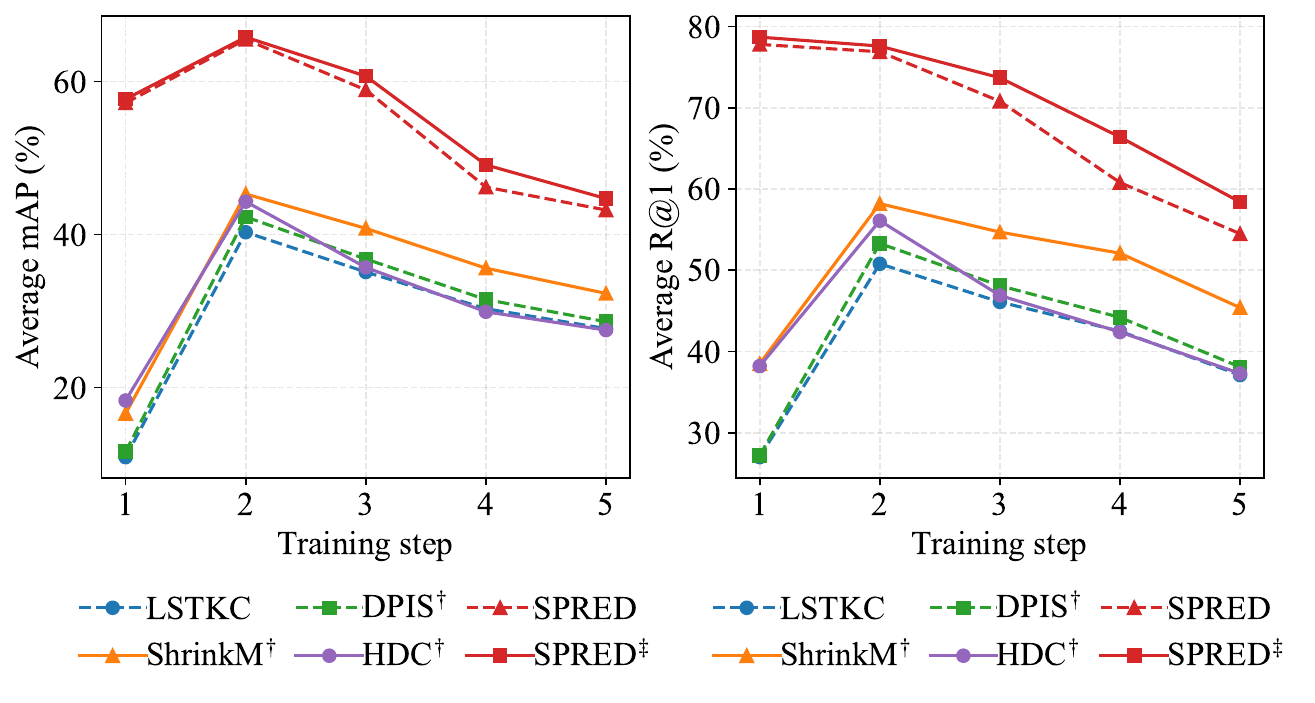}
    \vspace{-15pt}
        \caption{\label{fig:incremental} Seen domain performance curves under 10\% label rate. }
        \vspace{-5pt}
\end{figure}

\textbf{Seen Domain Performance Tendency}: In
\cref{fig:incremental}, 
we evaluate model performance on already encountered domains after each training step. The results indicate that both SPRED and SPRED${}^\ddag$ consistently outperform all existing methods.
% , demonstrating strong long-term knowledge accumulation capabilities. 
This advantage is attributed to our prototype-based anti-forgetting loss and effective pseudo-label purification mechanisms, which enhance knowledge consolidation and mitigate the catastrophic forgetting of historical correct knowledge caused by noisy pseudo-labels.

\subsection{Unseen-Domain Generalization Evaluation}
\textbf{Compared to LReID and SSL+LReID Methods}: 
As shown in Tab. \ref{tab:setting1}, our SPRED outperforms all LReID methods on the UnSeen-Avg mAP/R@1 performance. Besides, our SPRED${}^\ddag$ surpasses SSL+LReID approaches with \textbf{2.5\%/3.3\%}, \textbf{5.9\%/5.7\%}, and \textbf{7.9\%/7.5\%} UnSeen-Avg mAP/R@1 across label rate form 50\% to 10\%. These results underscore our method's effectiveness in accumulating generalizable knowledge, which is attributed to an efficient pseudo-label generation mechanism that reduces overfitting to the limited labeled data.

\begin{figure}[t]
    \centering
	\includegraphics[width=1\linewidth]{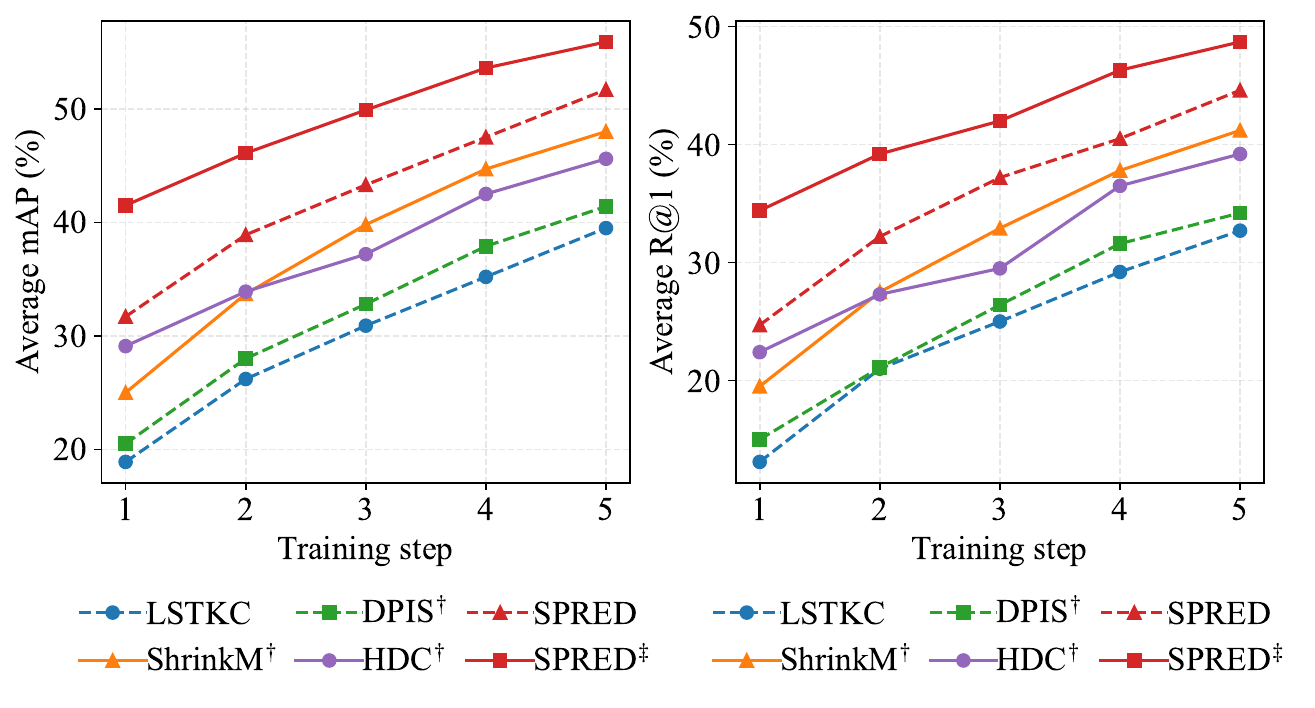}
        \vspace{-15pt}        \caption{\label{fig:generalization}Generalization curves under 10\% label rate.}
\end{figure}

\textbf{UnSeen Domain Performance Tendency}: In \cref{fig:generalization}, 
we assess the model’s performance on unseen domains after each training step. The results show that our SPRED and SPRED${}^\ddag$ achieve leading performance consistently, primarily due to enhanced utilization of unlabeled data, which integrates more generalizable knowledge into the model.
% Additionally, our prototype-oriented knowledge distillation strategy fortifies knowledge transfer, thereby improving the consolidation of generalizable knowledge.

\begin{table}[t!]
  \centering
  % \resizebox{.9\columnwidth}{!}{
  \setlength{\tabcolsep}{1.4mm}{
    \begin{tabular}{cccccccccc}
    \hline
    % \rowcolor{gray!30}
    && && \multicolumn{2}{c}{Seen-Avg}& \multicolumn{2}{c}{UnSeen-Avg} \\
    % \rowcolor{gray!30}
    \multirow{-1.9}[0]{*}{Base} & \multirow{-1.9}[0]{*}{DPL} & \multirow{-1.9}[0]{*}{NPL} & \multirow{-1.9}[0]{*}{DKCP}  & mAP   & R@1 &mAP   & R@1\\ 
          \hline
    $\checkmark$  &             &               & &27.7 &37.1 &39.5 &32.7 \\
    $\checkmark$  &$\checkmark$ &                & &28.8&38.2&41.7&34.2\\             
    $\checkmark$  &$\checkmark$ &$\checkmark$     &  &38.0&49.8&47.4&40.5 \\  
    $\checkmark$  &$\checkmark$ &$\checkmark$    &$\checkmark$ & \textbf{43.2} & \textbf{54.4}& \textbf{51.7} & \textbf{44.6} \\
    \hline
    \end{tabular}%
    \vspace{-10pt}
 }
    \caption{Ablation study of different components.}
  \label{tab:componment}%
\end{table}%
\begin{figure*}[t!]
		\centering
		\vspace{-2pt}  
            \subfloat[\label{fig:alpha} weight of $\mathcal{L}_s$]{
                \includegraphics[scale=0.345]{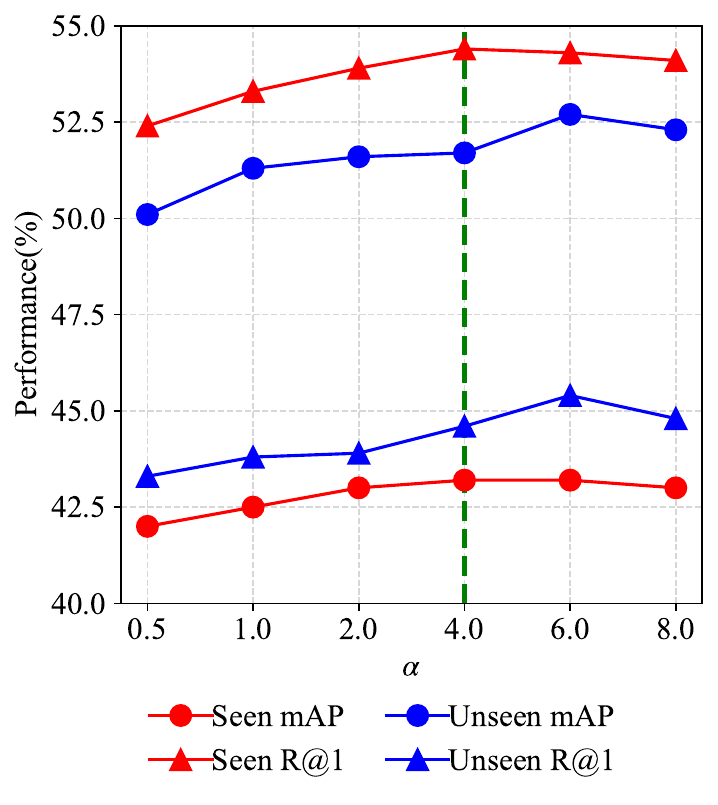}}
            % \hspace{1em} % 控制图像之间的间距
            \subfloat[\label{fig:T_c} new cluster label threshold $T_c$]{
                \includegraphics[scale=0.345]{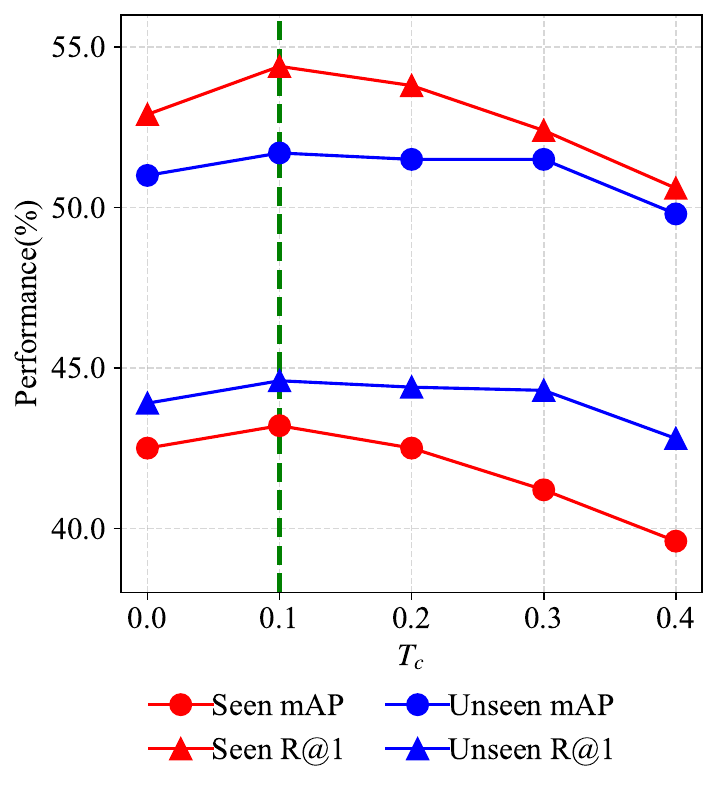}}
            % \hspace{1em} % 控制图像之间的间距
            \subfloat[\label{fig:proto-number} old cluster label threshold $T_o$]{
                \includegraphics[scale=0.345]{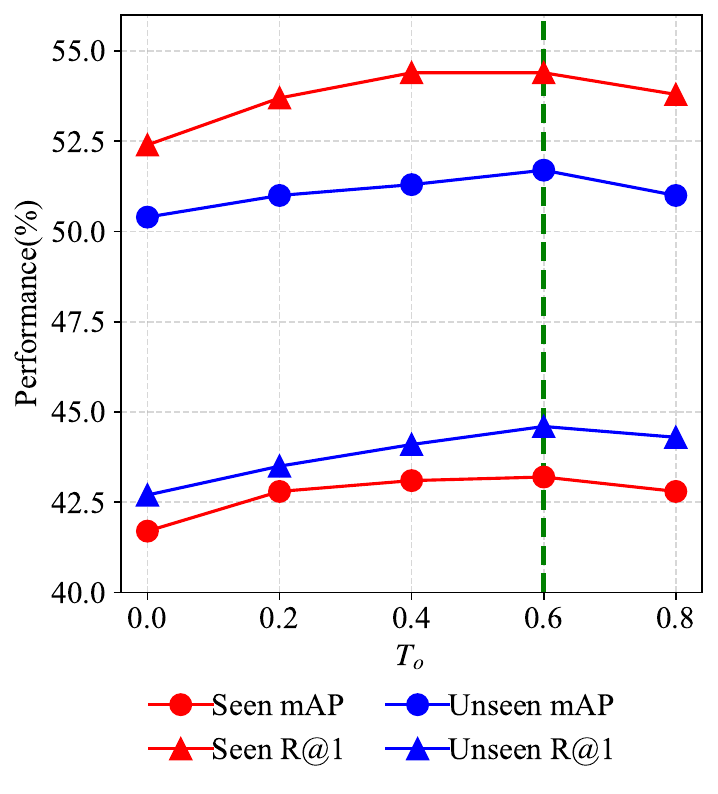}}
            % \hspace{1em} % 控制图像之间的间距
            \subfloat[\label{fig:sample-number} neighbor label threshold $T_p$]{
                \includegraphics[scale=0.345]{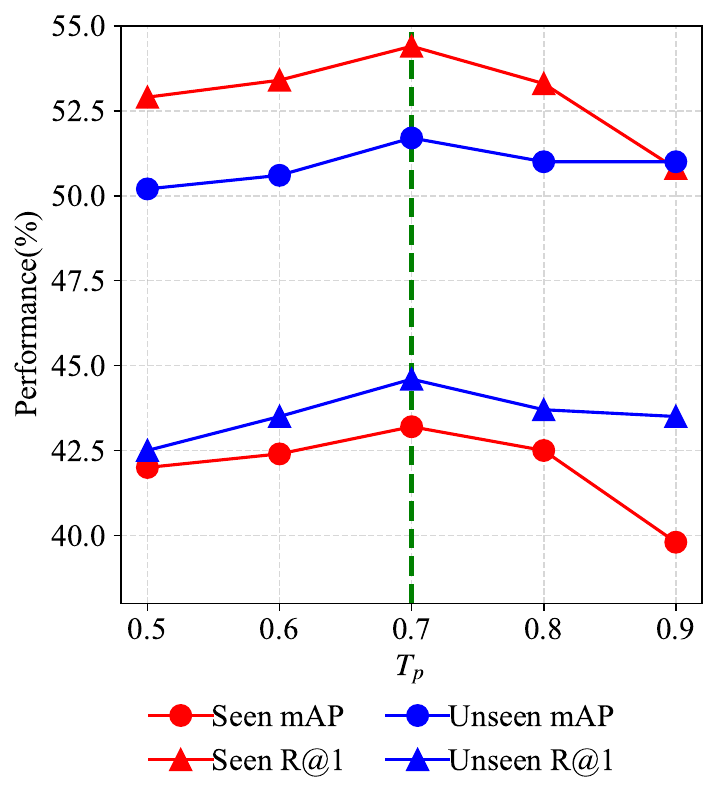}}
                \vspace{-5pt}  
		\caption{Ablation studies on hyperparameters under 10\% label rate. The default values are highlighted by the dashed lines.
		}
        \vspace{-5pt}  
		\label{fig:ablation} 
	\end{figure*}
 \subsection{Ablation Studies}
\textbf{Ablations on model components}.
	In ~\cref{tab:componment}, we adopt LSTKC as the Base model and progressively integrate the proposed DPL, NPL, and DKCP modules. The results show that each component substantially enhances model performance. When all components are incorporated, our method surpasses the Base model by 15.5\%/17.3\% Seen-Avg mAP/R@1 and 12.2\%/11.9\% UnSeen-Avg mAP/R@1. More ablations on the sub-components of DKCP are provided in our Supplementary Material.
    
	\textbf{Ablations on hyperparameters}. 
	We investigate the effects of the hyperparameters $\alpha$, $T_c$, $T_o$, and $T_p$, on the model in \cref{fig:ablation}. The results indicate that model performance is relatively robust to changes in these hyperparameters.
	In practice, we set $\alpha$, $T_c$, $T_o$, and $T_p$ to 4.0, 0.1, 0.6, and 0.7 by default, respectively.

 \begin{figure}[t]
 \vspace{-10pt}
    \centering
	\includegraphics[width=0.95\linewidth]{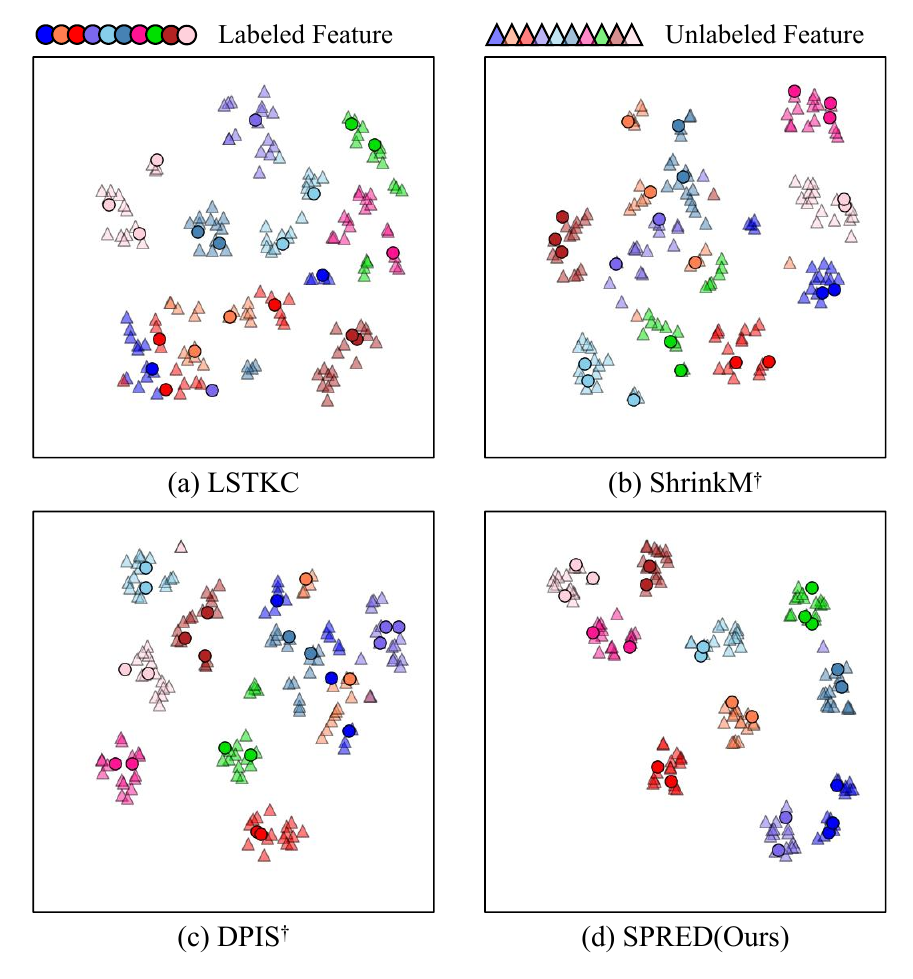}
    \vspace{-10pt}
        \caption{\label{fig:visualization}T-SNE visualization of labeled and unlabeled features in the training dataset MSMT17.}
        \vspace{-10pt}
\end{figure}

\subsection{Visualization Results}
\textbf{Feature Distribution Visualization}:
In \cref{fig:visualization}, 
we visualize the training data features of different identities (colors) in comparison with existing approaches. The results demonstrate that our method achieves a more cohesive grouping of labeled and unlabeled features while existing methods exhibit substantial labeled-unlabeled feature disparity and limited discriminative ability between identities. These results highlight our superiority in mining knowledge from unlabeled data and boosting the LReID performance.

\begin{figure}[t]
    \centering
    \vspace{-5pt}
	\includegraphics[width=0.96\linewidth]{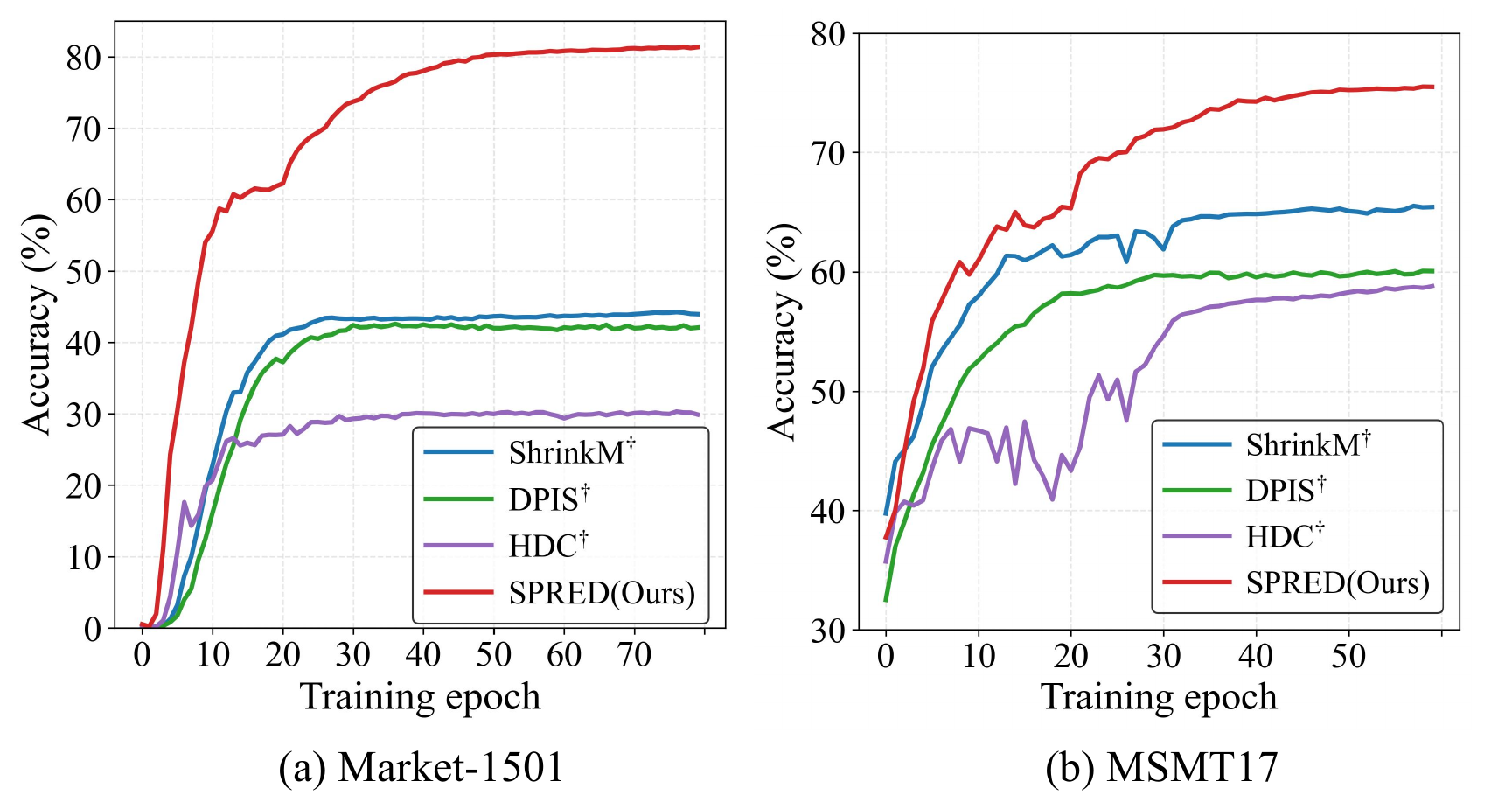}
        \vspace{-10pt}       
        \caption{\label{fig:label_acc} Label prediction accuracy of different methods.}
        \vspace{-10pt}  
\end{figure}

\textbf{Identity Prediction Visualization}: 
To demonstrate the knowledge acquisition capacity of different methods, we visualize the identity prediction accuracy on the challenging Market-1501 and MSMT17 datasets in \cref{fig:label_acc}. Unlike existing methods, whose accuracy plateaus after 25 epochs, our SPRED method continues to improve, ultimately achieving significantly higher final accuracy. These results confirm the effectiveness of our cyclically evolved unlabeled data utilization mechanism, in which pseudo-label prediction and purification iteratively enhance each other.

% To show the knowledge acquisition capacity of different methods when learning a new dataset, we visualize the identity prediction accuracy of different methods on challenging Market-1501 and MSMT17 datasets in \cref{fig:label_acc}.  Compared to the existing methods that stop to increase after 25epochs, our SPRED improves the accuracy continually and achieves significantly higher final accuracy. These results verify the effectiveness of our cyclicly evolved unlabeled data utilization mechanism where the pseudo-label prediction and purification facilitate each other in turn.
\vspace{-5pt}  
\section{Conclusion}

In this paper, a practical Semi-LReID task is investigated, aiming to learn from a data stream with a vast amount of unlabeled data alongside scarce labeled samples. To support further research, a new large-scale Semi-LReID benchmark is introduced. Additionally, a novel Semi-LReID approach, SPRED, is proposed. Specifically, a self-reinforcing cycle is established, wherein the learnable prototype is introduced to dynamically model identity distribution and generate high-quality pseudo-labels. Besides, a dual-knowledge collaborative purification mechanism is designed to filter the underlying unreliable pseudo-labels by exploiting the new knowledge and the complementary old knowledge. Extensive experiments demonstrate that the proposed method significantly outperforms existing approaches, particularly under conditions with limited labeled data.

%  \textbf{Acknowledgements}
% This work was supported by the National Key R\&D Program of China (2024YFA1410000) and the National Natural Science Foundation of China (62376011).

{
    \small
    \bibliographystyle{ieeenat_fullname}
    \bibliography{main}

\begin{thebibliography}{70}
\providecommand{\natexlab}[1]{#1}
\providecommand{\url}[1]{\texttt{#1}}
\expandafter\ifx\csname urlstyle\endcsname\relax
  \providecommand{\doi}[1]{doi: #1}\else
  \providecommand{\doi}{doi: \begingroup \urlstyle{rm}\Url}\fi

\bibitem[Berthelot et~al.(2019)Berthelot, Carlini, Goodfellow, Papernot, Oliver, and Raffel]{berthelot2019mixmatch}
David Berthelot, Nicholas Carlini, Ian Goodfellow, Nicolas Papernot, Avital Oliver, and Colin~A Raffel.
\newblock Mixmatch: A holistic approach to semi-supervised learning.
\newblock \emph{NeurIPS}, 32, 2019.

\bibitem[Brahma et~al.(2021)Brahma, Verma, and Rai]{brahma2021hypernetworks}
Dhanajit Brahma, Vinay~Kumar Verma, and Piyush Rai.
\newblock Hypernetworks for continual semi-supervised learning.
\newblock \emph{arXiv preprint arXiv:2110.01856}, 2021.

\bibitem[Branch(2006)]{branch2006imagery}
Home Office Scientific~Development Branch.
\newblock Imagery library for intelligent detection systems (i-lids).
\newblock In \emph{2006 IET conference on crime and security}, pages 445--448. IET, 2006.

\bibitem[Cai et~al.(2021)Cai, Ravichandran, Maji, Fowlkes, Tu, and Soatto]{cai2021exponential}
Zhaowei Cai, Avinash Ravichandran, Subhransu Maji, Charless Fowlkes, Zhuowen Tu, and Stefano Soatto.
\newblock Exponential moving average normalization for self-supervised and semi-supervised learning.
\newblock In \emph{CVPR}, pages 194--203. IEEE, 2021.

\bibitem[Chen et~al.(2023{\natexlab{a}})Chen, Guo, Wu, Qi, Li, and Dong]{chen2023learn}
Yongrui Chen, Xinnan Guo, Tongtong Wu, Guilin Qi, Yang Li, and Yang Dong.
\newblock Learn from yesterday: A semi-supervised continual learning method for supervision-limited text-to-sql task streams.
\newblock In \emph{AAAI}, pages 12682--12690, 2023{\natexlab{a}}.

\bibitem[Chen et~al.(2023{\natexlab{b}})Chen, Tan, Zhao, Chen, Song, Liang, and Lu]{chen2023boosting}
Yuhao Chen, Xin Tan, Borui Zhao, Zhaowei Chen, Renjie Song, Jiajun Liang, and Xuequan Lu.
\newblock Boosting semi-supervised learning by exploiting all unlabeled data.
\newblock In \emph{CVPR}, pages 7548--7557, 2023{\natexlab{b}}.

\bibitem[Cui et~al.(2024{\natexlab{a}})Cui, Zhou, and Peng]{cui2024dma}
Zhenyu Cui, Jiahuan Zhou, and Yuxin Peng.
\newblock Dma: Dual modality-aware alignment for visible-infrared person re-identification.
\newblock \emph{T-IFS}, 19:\penalty0 2696--2708, 2024{\natexlab{a}}.

\bibitem[Cui et~al.(2024{\natexlab{b}})Cui, Zhou, Wang, Zhu, and Peng]{cui2024learning}
Zhenyu Cui, Jiahuan Zhou, Xun Wang, Manyu Zhu, and Yuxin Peng.
\newblock Learning continual compatible representation for re-indexing free lifelong person re-identification.
\newblock In \emph{CVPR}, pages 16614--16623, 2024{\natexlab{b}}.

\bibitem[Cui et~al.(2025)Cui, Zhou, and Peng]{cui2025dkc}
Zhenyu Cui, Jiahuan Zhou, and Yuxin Peng.
\newblock Dkc: Differentiated knowledge consolidation for cloth-hybrid lifelong person re-identification.
\newblock In \emph{CVPR}, pages 3573--3582, 2025.

\bibitem[Fan et~al.(2024)Fan, Wang, Zhu, and Hu]{fan2024dynamic}
Yan Fan, Yu Wang, Pengfei Zhu, and Qinghua Hu.
\newblock Dynamic sub-graph distillation for robust semi-supervised continual learning.
\newblock In \emph{AAAI}, pages 11927--11935, 2024.

\bibitem[Fini et~al.(2023)Fini, Astolfi, Alahari, Alameda-Pineda, Mairal, Nabi, and Ricci]{fini2023semi}
Enrico Fini, Pietro Astolfi, Karteek Alahari, Xavier Alameda-Pineda, Julien Mairal, Moin Nabi, and Elisa Ricci.
\newblock Semi-supervised learning made simple with self-supervised clustering.
\newblock In \emph{CVPR}, pages 3187--3197, 2023.

\bibitem[Fu et~al.(2022)Fu, Du, Ding, Wang, Jiang, and Zhang]{fu2022domain}
Lihua Fu, Yubin Du, Yu Ding, Dan Wang, Hanxu Jiang, and Haitao Zhang.
\newblock Domain adaptive learning with multi-granularity features for unsupervised person re-identification.
\newblock \emph{Chinese Journal of Electronics}, 31\penalty0 (1):\penalty0 116--128, 2022.

\bibitem[Ge et~al.(2022)Ge, Du, Wu, Xian, Yan, Huang, and Zheng]{ge2022lifelong}
Wenhang Ge, Junlong Du, Ancong Wu, Yuqiao Xian, Ke Yan, Feiyue Huang, and Wei-Shi Zheng.
\newblock Lifelong person re-identification by pseudo task knowledge preservation.
\newblock In \emph{AAAI}, pages 688--696, 2022.

\bibitem[Gomez-Villa et~al.(2024)Gomez-Villa, Goswami, Wang, Bagdanov, Twardowski, and van~de Weijer]{gomez2024exemplar}
Alex Gomez-Villa, Dipam Goswami, Kai Wang, Andrew~D Bagdanov, Bartlomiej Twardowski, and Joost van~de Weijer.
\newblock Exemplar-free continual representation learning via learnable drift compensation.
\newblock In \emph{ECCV}, pages 473--490. Springer, 2024.

\bibitem[Gong et~al.()Gong, Zhong, Qu, Luo, Ji, and Jiang]{gongcross}
Yunpeng Gong, Zhun Zhong, Yansong Qu, Zhiming Luo, Rongrong Ji, and Min Jiang.
\newblock Cross-modality perturbation synergy attack for person re-identification.

\bibitem[Gong et~al.(2022)Gong, Huang, and Chen]{gong2022person}
Yunpeng Gong, Liqing Huang, and Lifei Chen.
\newblock Person re-identification method based on color attack and joint defence.
\newblock In \emph{CVPR}, pages 4313--4322, 2022.

\bibitem[Gray and Tao(2008)]{gray2008viewpoint}
Douglas Gray and Hai Tao.
\newblock Viewpoint invariant pedestrian recognition with an ensemble of localized features.
\newblock In \emph{ECCV}, pages 262--275. Springer, 2008.

\bibitem[Gu et~al.(2023)Gu, Luo, Wang, Jiang, You, and Zhao]{gu2023color}
Jianyang Gu, Hao Luo, Kai Wang, Wei Jiang, Yang You, and Jian Zhao.
\newblock Color prompting for data-free continual unsupervised domain adaptive person re-identification.
\newblock \emph{arXiv preprint arXiv:2308.10716}, 2023.

\bibitem[He et~al.(2021)He, Luo, Wang, Wang, Li, and Jiang]{he2021transreid}
Shuting He, Hao Luo, Pichao Wang, Fan Wang, Hao Li, and Wei Jiang.
\newblock Transreid: Transformer-based object re-identification.
\newblock In \emph{ICCV}, pages 14993--15002. IEEE, 2021.

\bibitem[Hershey and Olsen(2007)]{hershey2007approximating}
John~R Hershey and Peder~A Olsen.
\newblock Approximating the kullback leibler divergence between gaussian mixture models.
\newblock In \emph{ICASSP}, pages IV--317. IEEE, 2007.

\bibitem[Hirzer et~al.(2011)Hirzer, Beleznai, Roth, and Bischof]{hirzer2011person}
Martin Hirzer, Csaba Beleznai, Peter~M Roth, and Horst Bischof.
\newblock Person re-identification by descriptive and discriminative classification.
\newblock In \emph{Image Analysis}, pages 91--102. Springer, 2011.

\bibitem[Howard et~al.(2019)Howard, Sandler, Chu, Chen, Chen, Tan, Wang, Zhu, Pang, Vasudevan, et~al.]{howard2019searching}
Andrew Howard, Mark Sandler, Grace Chu, Liang-Chieh Chen, Bo Chen, Mingxing Tan, Weijun Wang, Yukun Zhu, Ruoming Pang, Vijay Vasudevan, et~al.
\newblock Searching for mobilenetv3.
\newblock In \emph{ICCV}, pages 1314--1324. IEEE, 2019.

\bibitem[Kang et~al.(2023)Kang, Fini, Nabi, Ricci, and Alahari]{kang2023soft}
Zhiqi Kang, Enrico Fini, Moin Nabi, Elisa Ricci, and Karteek Alahari.
\newblock A soft nearest-neighbor framework for continual semi-supervised learning.
\newblock In \emph{ICCV}, pages 11868--11877, 2023.

\bibitem[Lee and Kim(2024)]{lee2024cdmad}
Hyuck Lee and Heeyoung Kim.
\newblock Cdmad: Class-distribution-mismatch-aware debiasing for class-imbalanced semi-supervised learning.
\newblock In \emph{CVPR}, pages 23891--23900. IEEE, 2024.

\bibitem[Li et~al.(2021)Li, Xiong, and Hoi]{li2021comatch}
Junnan Li, Caiming Xiong, and Steven~CH Hoi.
\newblock Comatch: Semi-supervised learning with contrastive graph regularization.
\newblock In \emph{ICCV}, pages 9475--9484, 2021.

\bibitem[Li et~al.(2024)Li, Xu, Peng, and Zhou]{li2024exemplar}
Qiwei Li, Kunlun Xu, Yuxin Peng, and Jiahuan Zhou.
\newblock Exemplar-free lifelong person re-identification via prompt-guided adaptive knowledge consolidation.
\newblock \emph{IJCV}, pages 1--16, 2024.

\bibitem[Li and Wang(2013)]{li2013locally}
Wei Li and Xiaogang Wang.
\newblock Locally aligned feature transforms across views.
\newblock In \emph{CVPR}, pages 3594--3601. IEEE, 2013.

\bibitem[Li et~al.(2012)Li, Zhao, and Wang]{li2012human}
Wei Li, Rui Zhao, and Xiaogang Wang.
\newblock Human reidentification with transferred metric learning.
\newblock In \emph{ACCV}, pages 31--44. Springer, 2012.

\bibitem[Li et~al.(2014)Li, Zhao, Xiao, and Wang]{li2014deepreid}
Wei Li, Rui Zhao, Tong Xiao, and Xiaogang Wang.
\newblock Deepreid: Deep filter pairing neural network for person re-identification.
\newblock In \emph{CVPR}, pages 152--159. IEEE, 2014.

\bibitem[Lin et~al.(2019)Lin, Dong, Zheng, Yan, and Yang]{lin2019bottom}
Yutian Lin, Xuanyi Dong, Liang Zheng, Yan Yan, and Yi Yang.
\newblock A bottom-up clustering approach to unsupervised person re-identification.
\newblock In \emph{AAAI}, pages 8738--8745, 2019.

\bibitem[Liu and Tan(2021)]{liu2021certainty}
Lu Liu and Robby~T Tan.
\newblock Certainty driven consistency loss on multi-teacher networks for semi-supervised learning.
\newblock \emph{Pattern Recognition}, 120:\penalty0 108140, 2021.

\bibitem[Loy et~al.(2010)Loy, Xiang, and Gong]{loy2010time}
Chen~Change Loy, Tao Xiang, and Shaogang Gong.
\newblock Time-delayed correlation analysis for multi-camera activity understanding.
\newblock \emph{IJCV}, 90\penalty0 (1):\penalty0 106--129, 2010.

\bibitem[Luo et~al.(2024)Luo, Wong, Kankanhalli, and Zhao]{luo2024learning}
Yan Luo, Yongkang Wong, Mohan Kankanhalli, and Qi Zhao.
\newblock Learning to predict gradients for semi-supervised continual learning.
\newblock \emph{TNNLS}, 2024.

\bibitem[Pu et~al.(2021)Pu, Chen, Liu, Bakker, and Lew]{pu2021lifelong}
Nan Pu, Wei Chen, Yu Liu, Erwin~M Bakker, and Michael~S Lew.
\newblock Lifelong person re-identification via adaptive knowledge accumulation.
\newblock In \emph{CVPR}, pages 7897--7906. IEEE, 2021.

\bibitem[Pu et~al.(2022)Pu, Liu, Chen, Bakker, and Lew]{pu2022meta}
Nan Pu, Yu Liu, Wei Chen, Erwin~M Bakker, and Michael~S Lew.
\newblock Meta reconciliation normalization for lifelong person re-identification.
\newblock In \emph{ACM MM}, pages 541--549, 2022.

\bibitem[Pu et~al.(2023)Pu, Zhong, Sebe, and Lew]{pu2023memorizing}
Nan Pu, Zhun Zhong, Nicu Sebe, and Michael~S Lew.
\newblock A memorizing and generalizing framework for lifelong person re-identification.
\newblock \emph{IEEE TPAMI}, 45\penalty0 (11):\penalty0 13567--13585, 2023.

\bibitem[Sajjadi et~al.(2016)Sajjadi, Javanmardi, and Tasdizen]{sajjadi2016regularization}
Mehdi Sajjadi, Mehran Javanmardi, and Tolga Tasdizen.
\newblock Regularization with stochastic transformations and perturbations for deep semi-supervised learning.
\newblock \emph{NeurIPS}, 29, 2016.

\bibitem[Shi et~al.(2023)Shi, Zhang, Yin, Xie, Zhang, Fan, Shi, and Qu]{shi2023dual}
Jiangming Shi, Yachao Zhang, Xiangbo Yin, Yuan Xie, Zhizhong Zhang, Jianping Fan, Zhongchao Shi, and Yanyun Qu.
\newblock Dual pseudo-labels interactive self-training for semi-supervised visible-infrared person re-identification.
\newblock In \emph{ICCV}, pages 11218--11228. IEEE, 2023.

\bibitem[Shi et~al.(2024)Shi, Yin, Zhang, Xie, Qu, et~al.]{shi2024learning}
Jiangming Shi, Xiangbo Yin, Yachao Zhang, Yuan Xie, Yanyun Qu, et~al.
\newblock Learning commonality, divergence and variety for unsupervised visible-infrared person re-identification.
\newblock \emph{NeurIPS}, 37:\penalty0 99715--99734, 2024.

\bibitem[Shi et~al.(2025)Shi, Yin, Chen, Zhang, Zhang, Xie, and Qu]{shi2025multi}
Jiangming Shi, Xiangbo Yin, Yeyun Chen, Yachao Zhang, Zhizhong Zhang, Yuan Xie, and Yanyun Qu.
\newblock Multi-memory matching for unsupervised visible-infrared person re-identification.
\newblock In \emph{ECCV}, pages 456--474. Springer, 2025.

\bibitem[Smith et~al.(2021)Smith, Balloch, Hsu, and Kira]{smith2021memory}
James Smith, Jonathan Balloch, Yen-Chang Hsu, and Zsolt Kira.
\newblock Memory-efficient semi-supervised continual learning: The world is its own replay buffer.
\newblock In \emph{IJCNN}, pages 1--8. IEEE, 2021.

\bibitem[Sohn et~al.(2020)Sohn, Berthelot, Carlini, Zhang, Zhang, Raffel, Cubuk, Kurakin, and Li]{sohn2020fixmatch}
Kihyuk Sohn, David Berthelot, Nicholas Carlini, Zizhao Zhang, Han Zhang, Colin~A Raffel, Ekin~Dogus Cubuk, Alexey Kurakin, and Chun-Liang Li.
\newblock Fixmatch: Simplifying semi-supervised learning with consistency and confidence.
\newblock \emph{NeurIPS}, 33:\penalty0 596--608, 2020.

\bibitem[Song et~al.(2018)Song, Leng, Liu, Hetang, and Cai]{song2018region}
Guanglu Song, Biao Leng, Yu Liu, Congrui Hetang, and Shaofan Cai.
\newblock Region-based quality estimation network for large-scale person re-identification.
\newblock In \emph{AAAI}, 2018.

\bibitem[Sosea and Caragea(2023)]{sosea2023marginmatch}
Tiberiu Sosea and Cornelia Caragea.
\newblock Marginmatch: Improving semi-supervised learning with pseudo-margins.
\newblock In \emph{CVPR}, pages 15773--15782, 2023.

\bibitem[Sun and Mu(2022)]{sun2022patch}
Zhicheng Sun and Yadong Mu.
\newblock Patch-based knowledge distillation for lifelong person re-identification.
\newblock In \emph{ACM MM}, pages 696--707, 2022.

\bibitem[Wang et~al.(2021)Wang, Yang, Li, Hong, Li, and Zhu]{wang2021ordisco}
Liyuan Wang, Kuo Yang, Chongxuan Li, Lanqing Hong, Zhenguo Li, and Jun Zhu.
\newblock Ordisco: Effective and efficient usage of incremental unlabeled data for semi-supervised continual learning.
\newblock In \emph{CVPR}, pages 5383--5392, 2021.

\bibitem[Wang et~al.(2024)Wang, Zhu, Chen, and Hu]{wang2024persistence}
Yu Wang, Pengfei Zhu, Dongyue Chen, and Qinghua Hu.
\newblock Persistence homology distillation for semi-supervised continual learning.
\newblock \emph{NeurIPS}, 37:\penalty0 76332--76355, 2024.

\bibitem[Wei et~al.(2018)Wei, Zhang, Gao, and Tian]{wei2018person}
Longhui Wei, Shiliang Zhang, Wen Gao, and Qi Tian.
\newblock Person transfer gan to bridge domain gap for person re-identification.
\newblock In \emph{CVPR}, pages 79--88. IEEE, 2018.

\bibitem[Wei et~al.(2024)Wei, Yang, Wang, and Gao]{wei2024semi}
Ziyu Wei, Xi Yang, Nannan Wang, and Xinbo Gao.
\newblock Semi-supervised learning with heterogeneous distribution consistency for visible infrared person re-identification.
\newblock \emph{IEEE TIP}, 2024.

\bibitem[Wu et~al.(2023)Wu, Ge, and Zheng]{wu2023rewarded}
Ancong Wu, Wenhang Ge, and Wei-Shi Zheng.
\newblock Rewarded semi-supervised re-identification on identities rarely crossing camera views.
\newblock \emph{IEEE TPAMI}, 2023.

\bibitem[Wu and Gong(2021)]{wu2021generalising}
Guile Wu and Shaogang Gong.
\newblock Generalising without forgetting for lifelong person re-identification.
\newblock In \emph{AAAI}, pages 2889--2897, 2021.

\bibitem[Xiao et~al.(2016)Xiao, Li, Wang, Lin, and Wang]{xiao2016end}
Tong Xiao, Shuang Li, Bochao Wang, Liang Lin, and Xiaogang Wang.
\newblock End-to-end deep learning for person search.
\newblock \emph{arXiv preprint arXiv:1604.01850}, 2\penalty0 (2):\penalty0 4, 2016.

\bibitem[Xu et~al.(2024{\natexlab{a}})Xu, Zhang, Li, Peng, and Zhou]{xu2024mitigate}
Kunlun Xu, Haozhuo Zhang, Yu Li, Yuxin Peng, and Jiahuan Zhou.
\newblock Mitigate catastrophic remembering via continual knowledge purification for noisy lifelong person re-identification.
\newblock In \emph{ACM MM}, pages 1--9, 2024{\natexlab{a}}.

\bibitem[Xu et~al.(2024{\natexlab{b}})Xu, Zou, Peng, and Zhou]{xu2024distribution}
Kunlun Xu, Xu Zou, Yuxin Peng, and Jiahuan Zhou.
\newblock Distribution-aware knowledge prototyping for non-exemplar lifelong person re-identification.
\newblock In \emph{CVPR}, pages 16604--16613. IEEE, 2024{\natexlab{b}}.

\bibitem[Xu et~al.(2024{\natexlab{c}})Xu, Zou, and Zhou]{xu2024lstkc}
Kunlun Xu, Xu Zou, and Jiahuan Zhou.
\newblock Lstkc: Long short-term knowledge consolidation for lifelong person re-identification.
\newblock In \emph{AAAI}, pages 16202--16210, 2024{\natexlab{c}}.

\bibitem[Xu et~al.(2025{\natexlab{a}})Xu, Jiang, Xiong, Peng, and Zhou]{xu2025dask}
Kunlun Xu, Chenghao Jiang, Peixi Xiong, Yuxin Peng, and Jiahuan Zhou.
\newblock Dask: Distribution rehearsing via adaptive style kernel learning for exemplar-free lifelong person re-identification.
\newblock In \emph{AAAI}, pages 8915--8923, 2025{\natexlab{a}}.

\bibitem[Xu et~al.(2025{\natexlab{b}})Xu, Liu, Zou, Peng, and Zhou]{xu2025long}
Kunlun Xu, Zichen Liu, Xu Zou, Yuxin Peng, and Jiahuan Zhou.
\newblock Long short-term knowledge decomposition and consolidation for lifelong person re-identification.
\newblock \emph{IEEE TPAMI}, 2025{\natexlab{b}}.

\bibitem[Xu et~al.(2025{\natexlab{c}})Xu, Zou, Hua, and Zhou]{xu2025componential}
Kunlun Xu, Xu Zou, Gang Hua, and Jiahuan Zhou.
\newblock Componential prompt-knowledge alignment for domain incremental learning.
\newblock In \emph{ICML}, 2025{\natexlab{c}}.

\bibitem[Yang et~al.(2023{\natexlab{a}})Yang, Zhao, Qi, Qiao, Shi, and Zhao]{yang2023shrinking}
Lihe Yang, Zhen Zhao, Lei Qi, Yu Qiao, Yinghuan Shi, and Hengshuang Zhao.
\newblock Shrinking class space for enhanced certainty in semi-supervised learning.
\newblock In \emph{ICCV}, pages 16187--16196, 2023{\natexlab{a}}.

\bibitem[Yang et~al.(2023{\natexlab{b}})Yang, Wu, Zhang, Li, and Wang]{yang2023handling}
Zexian Yang, Dayan Wu, Wanqian Zhang, Bo Li, and Weipinng Wang.
\newblock Handling label uncertainty for camera incremental person re-identification.
\newblock In \emph{ACM MM}, page 6253–6263, 2023{\natexlab{b}}.

\bibitem[Ye et~al.(2022)Ye, He, and Peng]{ye2022unsupervised}
Zhaoda Ye, Xiangteng He, and Yuxin Peng.
\newblock Unsupervised cross-media hashing learning via knowledge graph.
\newblock \emph{Chinese Journal of Electronics}, 31\penalty0 (6):\penalty0 1081--1091, 2022.

\bibitem[Yin et~al.(2024)Yin, Shi, Zhang, Lu, Zhang, Xie, and Qu]{yin2024robust}
Xiangbo Yin, Jiangming Shi, Yachao Zhang, Yang Lu, Zhizhong Zhang, Yuan Xie, and Yanyun Qu.
\newblock Robust pseudo-label learning with neighbor relation for unsupervised visible-infrared person re-identification.
\newblock In \emph{ACM MM}, pages 2242--2251, 2024.

\bibitem[Yu et~al.(2023)Yu, Shi, Liu, Gao, and Wang]{yu2023lifelong}
Chunlin Yu, Ye Shi, Zimo Liu, Shenghua Gao, and Jingya Wang.
\newblock Lifelong person re-identification via knowledge refreshing and consolidation.
\newblock In \emph{AAAI}, pages 3295--3303, 2023.

\bibitem[Zhai et~al.(2019)Zhai, Oliver, Kolesnikov, and Beyer]{zhai2019s4l}
Xiaohua Zhai, Avital Oliver, Alexander Kolesnikov, and Lucas Beyer.
\newblock S4l: Self-supervised semi-supervised learning.
\newblock In \emph{ICCV}, pages 1476--1485, 2019.

\bibitem[Zhang et~al.(2025)Zhang, Xu, Liu, Peng, and Zhou]{zhang2025scap}
Chenyu Zhang, Kunlun Xu, Zichen Liu, Yuxin Peng, and Jiahuan Zhou.
\newblock Scap: Transductive test-time adaptation via supportive clique-based attribute prompting.
\newblock In \emph{CVPR}, pages 30032--30041, 2025.

\bibitem[Zhao et~al.(2017)Zhao, Tian, Sun, Shao, Yan, Yi, Wang, and Tang]{zhao2017spindle}
Haiyu Zhao, Maoqing Tian, Shuyang Sun, Jing Shao, Junjie Yan, Shuai Yi, Xiaogang Wang, and Xiaoou Tang.
\newblock Spindle net: Person re-identification with human body region guided feature decomposition and fusion.
\newblock In \emph{CVPR}, pages 907--915. IEEE, 2017.

\bibitem[Zheng et~al.(2015)Zheng, Shen, Tian, Wang, Wang, and Tian]{zheng2015scalable}
Liang Zheng, Liyue Shen, Lu Tian, Shengjin Wang, Jingdong Wang, and Qi Tian.
\newblock Scalable person re-identification: A benchmark.
\newblock In \emph{ICCV}, pages 1116--1124. IEEE, 2015.

\bibitem[Zheng et~al.(2022)Zheng, You, Huang, Wang, Qian, and Xu]{zheng2022simmatch}
Mingkai Zheng, Shan You, Lang Huang, Fei Wang, Chen Qian, and Chang Xu.
\newblock Simmatch: Semi-supervised learning with similarity matching.
\newblock In \emph{CVPR}, pages 14471--14481, 2022.

\bibitem[Zheng et~al.(2023)Zheng, You, Huang, Luo, Wang, Qian, and Xu]{zheng2023simmatchv2}
Mingkai Zheng, Shan You, Lang Huang, Chen Luo, Fei Wang, Chen Qian, and Chang Xu.
\newblock Simmatchv2: Semi-supervised learning with graph consistency.
\newblock In \emph{CVPR}, pages 16432--16442. IEEE, 2023.

\bibitem[Zhong et~al.(2018)Zhong, Zheng, Zheng, Li, and Yang]{zhong2018camera}
Zhun Zhong, Liang Zheng, Zhedong Zheng, Shaozi Li, and Yi Yang.
\newblock Camera style adaptation for person re-identification.
\newblock In \emph{CVPR}, pages 5157--5166. IEEE, 2018.

\end{thebibliography}
}

% ICCV 2025 Paper Template
\clearpage

\maketitlesupplementary

In our supplementary materials, we provide additional implementation details for the proposed approach and the Semi-LReID benchmark. Furthermore, we include extensive quantitative and qualitative results that emphasize the effectiveness of our method in comparison to existing works.
In summary, the supplementary materials primarily include:
\begin{itemize}
\item Detailed implementation and architecture of the proposed distribution alignment network (DANet).
\item Ablation studies on the components of the DKCP module.
\item Pseudo-code of our SPRED method.
\item Overview of the ReID datasets and their arrangement in the proposed Semi-LReID benchmark.
\item Quantitative comparisons with state-of-the-art methods under different dataset orders.
\item Quantitative comparisons with state-of-the-art methods at a label rate of $r$=100\%.
\item Quantitative comparisons with state-of-the-art NoisyLReID and SSLL methods.
\item Qualitative results on the feature distribution and label prediction accuracy.
\end{itemize}

\section{Implementation of Distribution Alignment Network}
Inspired by \cite{gu2023color}, which revealed that the primary distribution difference across ReID datasets lies in color styles, we propose a distribution alignment network (DANet) based on the architecture of \cite{gu2023color}.

\begin{figure}[t]
		\begin{center}
			\includegraphics[width=1.0\linewidth]{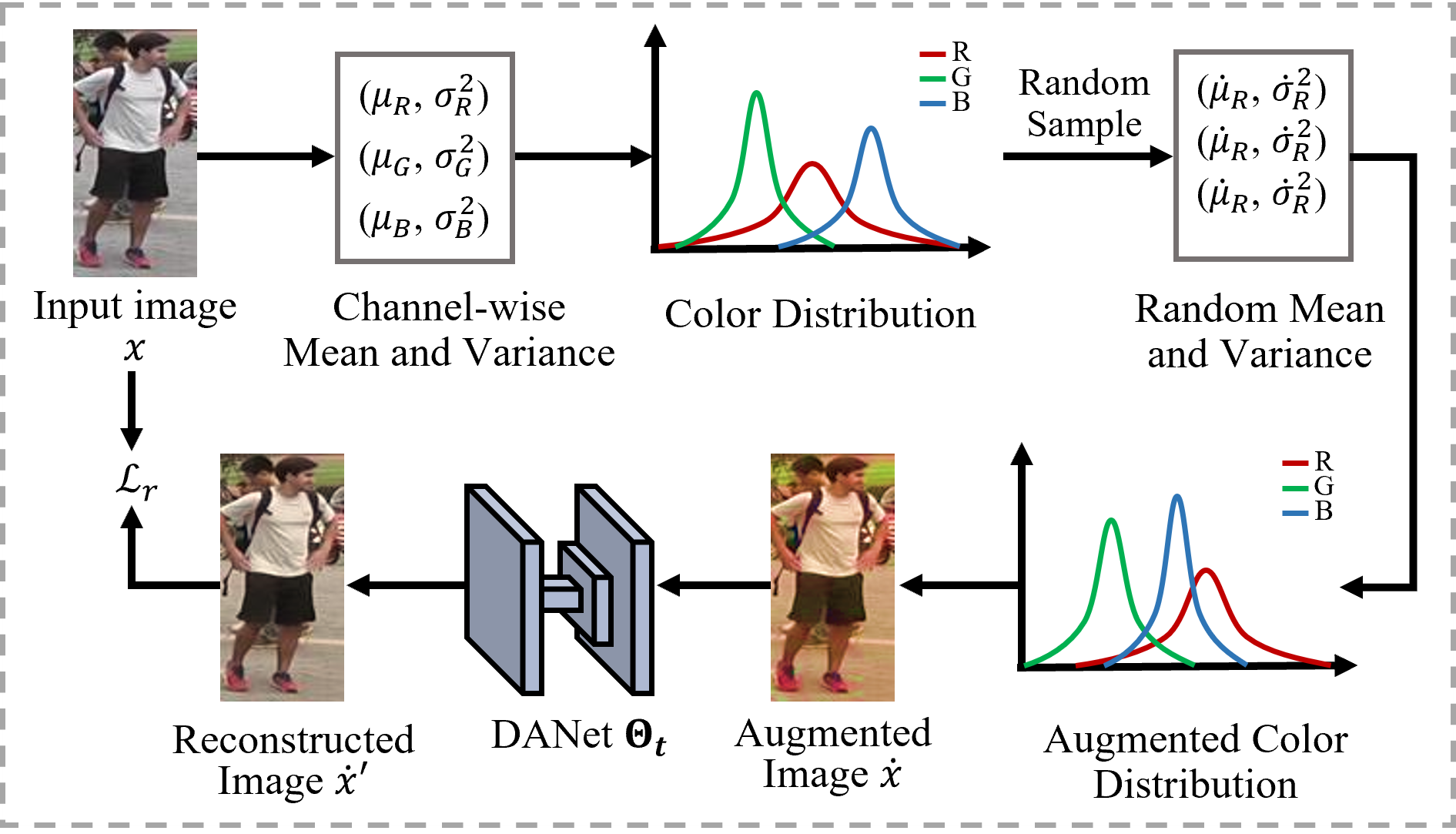}
			\caption{Training process of distribution alignment network (DANet)}
			\label{fig:DANet}
		\end{center}    
\end{figure}

As illustrated in~\cref{fig:DANet}, for a given image $x$, the channel-wise mean and variance are calculated, \ie,
 ($\mu_{R}$, $\sigma_{R}$), ($\mu_{G}$, and $\sigma_{G}$), and ($\mu_{B}$, $\sigma_{B}$). These statistics are modeled as Gaussian distributions per channel. For example, the $R$-channel mean and standard deviation follow $\mathcal{N}(\mu_R, \sigma^2_R)$ and $\mathcal{N}(\sigma_R, \sigma^2_R)$, respectively. Next, the color augmentation is conducted which involves sampling new mean and standard deviation values for each channel. For instance, for the $R$-channel, given a sampled mean $\dot{\mu}_R$ and standard deviation $\dot{\sigma}_R$, the augmented image $\dot{x}_R$ is computed as: 
 \begin{equation}
   \dot{x}_R=(x_R-\mu_R+\dot{\mu}_R)\frac{\dot{\sigma}_R}{\sigma_R}
\end{equation}
   Similarly, the augmented $G$ channel $\dot{x}_G$ and $B$ channel $\dot{x}_B$ are generated, forming the final augmented image $\dot{x}$.

Then, we introduce a lightweight distribution alignment network (DANet) $\boldsymbol{\Theta}_t$ which contains a MobileNet \cite{howard2019searching} backbone and a convolutional decoder. $\boldsymbol{\Theta}_t$ generates a reconstructed image $\dot{x}'$ from $\dot{x}$. The reconstruction process is guided by a mean absolute error (MAE) loss $\mathcal{L}_{r}$, defined as:
 \begin{equation}
   \mathcal{L}_{r}=||x-\dot{x}'||
    \label{eq:loss-reconstruct}
\end{equation}

Different from \cite{gu2023color}, which employed additional LBP transformations and formed a multi-step process for reconstruction during testing, our approach performs input-wise augmentation and direct reconstruction. This ensures an efficient end-to-end workflow, significantly simplifying applications while maintaining robust alignment.

Note that during the $t$-th ($t > 1$) training step, the previous DANet $\boldsymbol{\Theta}{t-1}$ is utilized to improve the adaptation of old knowledge to the new dataset $D_t$. Additionally, $\boldsymbol{\Theta}{t-1}$ only needs to forward once for each image, and the generated clustering results from the Old Knowledge Aligning-based Clustering (OKAC) strategy can be reused across training epochs, ensuring computational efficiency.

To verify the effectiveness of distribution alignment, we conduct an ablation study in \cref{tab:ablation-DANet}. The results show that when introducing $\boldsymbol{\Theta}_{t-1}$, consistent \textbf{0.6\%-1.0\%} improvement on Seen-Avg and UnSeen-Avg performance is achieved in our framework.

\section{Ablation Studies on DKCP}
Our Dual-Knowledge Cooperation-driven Pseudo-label Purification (DKCP) module is designed to filter noisy pseudo-labels by leveraging both old and new model knowledge. To achieve this,  two complementary strategies are introduced: Old Knowledge Aligning-based Clustering (OKAC) and New Knowledge Clustering (NKC), which extract knowledge from the old and new models, respectively.

To verify the effectiveness of our dual-knowledge utilization, we conduct ablation studies as summarized in \cref{tab:ablation-DKCP}. Starting from the DPL+NPL without DKCP, we progressively incorporate OKAC and NKC.
The results demonstrate that integrating either OKAC or NKC individually improves Semi-LReID performance. This improvement occurs because prototype-based pseudo-label generation primarily focuses on prototype-instance relations, neglecting inter-instance relations. By contrast, OKAC and NKC extract inter-instance relations through clustering, complementing the prototype-based pseudo-label generation. 
When both OKAC and NKC are utilized together, the performance is further enhanced. This advancement arises from our method effectively combining the generalizable old knowledge and abundant new knowledge, leading to the generation of higher-quality pseudo-labels. 

\begin{table}[ht]
  \centering
  % \resizebox{.9\columnwidth}{!}{
  \setlength{\tabcolsep}{1.2mm}{
    \begin{tabular}{lccccccccc}
    \hline
    % \rowcolor{gray!30}
    & \multicolumn{2}{c}{Seen-Avg}& \multicolumn{2}{c}{UnSeen-Avg} \\
    % \rowcolor{gray!30}
    \multicolumn{1}{c}{\multirow{-1.9}[0]{*}{Method}} &  mAP   & R@1 &mAP   & R@1\\ 
          \hline
    Ours (w/o DANet $\boldsymbol{\Theta}_{t-1}$)  &42.5 &53.8 &51.1 &43.6\\
    Ours (w/ DANet $\boldsymbol{\Theta}_{t-1}$)  & \textbf{43.2} & \textbf{54.4}& \textbf{51.7} & \textbf{44.6} \\
    % 43.2 54.4 51.7 44.6
    \hline
    \end{tabular}%
 }
    \caption{Ablation study on DANet $\boldsymbol{\Theta}_{t-1}$.}
  \label{tab:ablation-DANet}%
\end{table}%

\begin{table*}[ht]
  \centering  
  \setlength{\tabcolsep}{5.4mm}{
    \begin{tabular}{l|l|llc|ccc}
    \hline
    \multirow{2}[0]{*}{Type} & \multirow{2}[0]{*}{Datasets Name} & \multicolumn{3}{c|}{Original Identities} & \multicolumn{3}{c}{Semi-LReID Identities} \\
\cline{3-8}          &       & Train & Query & Gallery & Train & Query & Gallery \\
    \hline
    \multirow{5}[2]{*}{Seen} & CUHK03~\cite{li2014deepreid} & 767   & 700   & 700   & 500   & 700   & 700 \\
          & Market-1501~\cite{zheng2015scalable} & 751   & 750   & 751   & 500   & 751   & 751 \\
          & LPW~\cite{song2018region} & 875   & 876   & 876  & 500   & 876   & 876 \\
          & CUHK-SYSU~\cite{xiao2016end} & 942   & 2900  & 2900  & 500   & 2900  & 2900 \\
          & MSMT17-V2~\cite{wei2018person}& 1041  & 3060  & 3060  & 500   & 3060  & 3060 \\          
    \hline
    \multirow{7}[2]{*}{Unseen}
    & i-LIDS~\cite{branch2006imagery}& 243   & 60    & 60    & -      & 60    & 60 \\
    & VIPR~\cite{gray2008viewpoint} & 316   & 316   & 316   &  -     & 316   & 316 \\
    & GRID~\cite{loy2010time} & 125   & 125   & 126   &  -     & 125   & 126 \\   
          & PRID~\cite{hirzer2011person}& 100   & 100   & 649   & -      & 100   & 649 \\
                 
          & CUHK01~\cite{li2012human} & 485   & 486   & 486   & -      & 486   & 486 \\
          & CUHK02~\cite{li2013locally} & 1677  & 239   & 239   & -      & 239   & 239 \\
          & SenseReID~\cite{zhao2017spindle}& 1718  & 521   & 1718  & -      & 521   & 1718 \\
    \hline
    \end{tabular}%
    }
    \caption{The statistics of datasets used in the Semi-LReID benchmark. ‘-’ indicates the dataset is only used as a test domain.}   
    \label{tab:datasets}%
\end{table*}%

\begin{table}[htbp]
  \centering
  % \vspace{-3pt}
  \setlength{\tabcolsep}{1.4mm}{
    \begin{tabular}{cccccccccc}
    \hline
    % \rowcolor{gray!30}
    && & \multicolumn{2}{c}{Seen-Avg}& \multicolumn{2}{c}{UnSeen-Avg} \\
    % \rowcolor{gray!30}
    \multirow{-1.9}[0]{*}{DPL+NPL} &  \multirow{-1.9}[0]{*}{OKAC} & \multirow{-1.9}[0]{*}{NKC}  & mAP   & R@1 &mAP   & R@1\\ 
          \hline
    $\checkmark$  &             &               &38.0&49.8&47.4&40.5\\
    $\checkmark$  &$\checkmark$ &               &39.6&51.2&48.3&42.8\\             
    $\checkmark$  &             &$\checkmark$   &40.1&51.4 &47.6&41.0\\  
    $\checkmark$  &$\checkmark$ &$\checkmark$   &  \textbf{43.2} & \textbf{54.4}& \textbf{51.7} & \textbf{44.6}\\
    \hline
    % \vspace{-3pt}
    \end{tabular}%
 }
    \caption{Ablation study on components of DKCP. `DPL+NPL' represents our method without DKCP.}
    % \vspace{-3pt}
  \label{tab:ablation-DKCP}%
\end{table}%

\section{Algorithm}
The pseudo-code of our SPRED method is shown in Alg~\ref{alg:alg1}.
\begin{algorithm}[h!]
   \caption{SPRED Algorithm}
   \label{alg:alg1}
\begin{algorithmic}
   \STATE {\bfseries Input:}  $D_t=\{(X_t^l,Y_t^l)\}\cup\{X_t^u\}$, $\boldsymbol{\mathrm{M}}_{t-1}$, $\mathcal{P}_{t-1}=\{p_{t-1}^i\}_{i=1}^{L_{t-1}}$.
   \STATE {\bfseries Output:} $\boldsymbol{\mathrm{M}}_{t}$, $\mathcal{P}_{t}=\{p_{t}^i\}_{i=1}^{L_{t}}$
   \STATE   
   \STATE \textit{\# Old Knowledge Aligning-based Clustering} 
   \STATE $D_t^*=$DANet$(D_t)$; \textit{\quad \# Apply style transfer} 
   \STATE $F_{t-1}^*=\boldsymbol{\mathrm{M}}_{t-1}(D_t^*)$; \textit{\quad \# Extract feature} 
   \STATE $\mathcal{C}_{t-1} = \{C_1^{t-1}, C_2^{t-1}, \dots, C_{N_C^{t-1}}^{t-1}\}$; \textit{\quad\# Apply clustering}
   \STATE
% $(X_t^{pse},Y_t^{pse})=(X_t^l,Y_t^l)$, 
   \STATE Initialize $\boldsymbol{\mathrm{M}}_{t}^0=\boldsymbol{\mathrm{M}}_{t-1}$
   \FOR{$e=1$ {\bfseries to} $N_{epoch}$}
   \STATE \textit{\# Neighbor Prototype Labeling}
   \FOR {$x$ {\bfseries in} $X_t^u$}
        \STATE $f_t^{e-1}=\boldsymbol{\mathrm{M}}_{t}^{e-1}(x)$;
        \STATE Obtain top-2 nearest prototypes $p_A$ and $p_B$;
        \STATE $s_A=\frac{e^{<f_t^{e-1}, p_A>}}{e^{<f_t^{e-1}, p_A>}+e^{<f_t^{e-1}, p_B>}}$;\textit{\#Classification score}
   \STATE Obtain $l_i$ according to Eq. (6);
    \ENDFOR 
   
   \STATE
   \STATE \textit{\# New Knowledge Clustering}
    \STATE $F_{t}^{e-1}=\boldsymbol{\mathrm{M}}_{t}^{e-1}(D_t)$; \textit{\quad \# Extract feature} 
   \STATE $\mathcal{C}_{t}^{e-1}=\{C_1^{e-1},C_2^{e-1},...,C_{N_C^{t,e-1}}^{e-1}\}$; \textit{\quad\# Clustering}
   \STATE $ Y_t^{pse}=Y_{t-1}^{pse} \cup Y_{e-1}^{pse}$;
   \STATE
   \STATE \textit{\# Noisy Pseudo-Label Filtering}
   \STATE Obtain $Y_{e-1}^{pse}=\{y^{pse}_{i,e-1}: x_i\in X_t^u\}$ according to $LC_s(x_i,\mathcal{S}^{e-1}, \mathcal{C}_t^{e-1})$;
   \STATE Obtain $Y_{t-1}^{pse}=\{y^{pse}_{i,t-1}: x_i\in X_t^u\}$ according to $LC_s(x_i,\mathcal{S}^{e-1}, \mathcal{C}_{t-1})$;
   \STATE
    \STATE \textit{\# Dynamic Prototype-guided LReID Learning}
    \FOR {$(x,y)$ {\bfseries in} $(X_t^{pse},Y_t^{pse})$}
        \STATE $\mathcal{L}_{base}=\mathcal{L}_{p}+\mathcal{L}_{Tri}$, \STATE $\mathcal{L}_{s}=\mathcal{L}_{KL}\big(S(f_{t-1}$
        $\mathcal{P}_{t-1})||S(f_{t}^e, \mathcal{P}_{t-1})\big)$;
        \STATE Optimize $\boldsymbol{\mathrm{M}}_{t}^e$ with loss $\mathcal{L}=\mathcal{L}_{base} +\alpha \mathcal{L}_{s}$;        
    \ENDFOR

   \ENDFOR 
   \STATE $\boldsymbol{\mathrm{M}}_{t}=\delta_t\boldsymbol{\mathrm{M}}_{t-1}+(1-\delta_t)\boldsymbol{\mathrm{M}}_{t}^{N_{epoch}}$;\textit{\quad\# $\delta_t$ is obtained according to LSTKC} \cite{xu2024lstkc}
   \STATE Return $\boldsymbol{\mathrm{M}}_{t}$, $\mathcal{P}_{t}=\{p_{t}^i\}_{i=1}^{L_{t}}$;
      
\end{algorithmic}
\end{algorithm}

\section{Datasets Details of Semi-LReID Benchmark}
We establish the Semi-LReID benchmark based on the existing LReID configuration\cite{pu2021lifelong}, incorporating 12 widely-used ReID datasets: Market1501~\cite{zheng2015scalable}, LPW~\cite{song2018region}, CUHK-SYSU~\cite{xiao2016end}, MSMT17-V2~\cite{wei2018person},  CUHK03 \cite{li2014deepreid}, CUHK01~\cite{li2012human}, CUHK02~\cite{li2013locally}, VIPeR~\cite{gray2008viewpoint}, PRID~\cite{hirzer2011person}, i-LIDS~\cite{branch2006imagery}, GRID~\cite{loy2010time}, and SenseReID~\cite{zhao2017spindle}). The detailed dataset statistics are presented in \cref{tab:datasets}. 
Among these datasets, CUHK-SYSU is initially proposed for the person search. To adapt it for ReID, we crop individual-level images using the ground-truth bounding box annotations. Then, a subset where each identity has at least 4 bounding boxes is selected following \cite{pu2021lifelong}.
Additionally, LPW~\cite{song2018region}, originally a video-based person re-identification dataset, is transformed into a ReID dataset by sampling one frame every 15 frames from the original sequences. The resulting dataset is then structured in accordance with the Market1501~\cite{zheng2015scalable} format.
 In addition, to mitigate the data imbalance between datasets~\cite{pu2021lifelong,pu2023memorizing}, 500 identities of each dataset are selected to form the Semi-LReID benchmark. In \cref{tab:datasets}, the column `Original Identities' represents the total number of identities in each dataset, while `Semi-LReID Identities' lists the selected identities for our benchmark.
To generate training data under different label rates, we ensure that each identity includes at least two labeled samples. Random selection is then applied to the remaining data until the predefined label rate is satisfied. 
 % This design avoids the serious label number imbalance across datasets by introducing minimal label labor.
\begin{table*}[ht]
    \centering          
          \setlength{\tabcolsep}{0.9mm}{
            \begin{tabular}{cclcccccccccc>{\columncolor{seen_back}}c>{\columncolor{seen_back}}c>{\columncolor{unseen_back}}c>{\columncolor{unseen_back}}cccc}   

           	\hline
                % \multirow{2}[1]{*}{\rotatebox{90}{\makecell{Label\\ Rate}}} 
                \multirow{2}[1]{*}{$r$} 
                % & \multirow{2}[1]{*}{\rotatebox{90}{Type}} 
            & \multirow{2}[1]{*}{Type} 
            & \multirow{2}[1]{*}{Method} &             
            \multicolumn{2}{c}{LPW} & \multicolumn{2}{c}{MSMT17} & \multicolumn{2}{c}{Market-1501} & \multicolumn{2}{c}{CUHK-SYSU} & 
            \multicolumn{2}{c}{CUHK03} & \multicolumn{2}{>{\columncolor{seen_back}}c}{\textbf{Seen-Avg}} & 
                \multicolumn{2}{>{\columncolor{unseen_back}}c}{\textbf{UnSeen-Avg}} \\

                \hhline{~~~----------------}
				% \cline{4-17} 
                & &   & mAP & R@1  & mAP  & R@1  & mAP  & R@1  & mAP  & R@1  & mAP  & R@1  & mAP  & R@1  & mAP  & R@1 \\    
				\hline
				\multirow{11}[1]{*}{\rotatebox{90}{50\%}} &\multirow{3.3}[0]{*}{\rotatebox{90}{LReID}} &
                PatchKD~\cite{sun2022patch}  &45.6	&57.6	&4.2	&11.9	&34.4	&58.7	&70.6	&74.0	&19.5	&19.6	&34.9	&44.4	&38.5	&33.4
 \\		
                % & & CKP~\cite{xu2024lstkc}  &55.2  &78.5  &81.0  &82.8  &44.3  &56.9  &17.6  &40.5  &39.2  &40.9  &47.5  &59.9  &55.7  &48.1  \\
				& & DKP~\cite{xu2024distribution}    &41.0	&53.2	&12.2	&29.8	&49.8	&72.7	&81.5	&83.7	&27.7	&26.9	&42.4	&53.3	&53.8	&46.9
 \\
				& & LSTKC~\cite{xu2024lstkc} &37.2	&48.7	&13.1	&31.8	&51.9	&73.4	&81.1	&83.2	&37.9	&37.9	&44.2	&55.0	&53.3	&46.0

  \\
				% \cline{2-17}
                \hhline{~----------------}
				&\multirow{5.3}[0]{*}{\rotatebox{90}{SSL+LReID}} & 
    
				CDMAD${}^\dag$~\cite{lee2024cdmad} &30.4	&45.7	&6.9	&22.5	&27.9	&57.5	&76.2	&79.8	&11.9	&11.0	&30.7	&43.3	&46.8	&40.7

\\
				&& ShrinkM${}^\dag$\cite{yang2023shrinking}    &40.3	&54.8	&18.8	&44.8	&53.5	&77.0	&77.5	&79.7	&36.4	&36.8	&45.3	&58.6	&\textcolor{blue}{\textbf{57.7}}	&\textcolor{blue}{\textbf{50.9}}

 \\
                &&SimMV2${}^\dag$~\cite{zheng2023simmatchv2}  &27.3	&39.0	&8.7	&25.3	&42.4	&66.8	&63.0	&66.1	&29.7	&32.4	&34.2	&45.9	&44.8	&38.0

 \\
                && DPIS${}^\dag$\cite{shi2023dual}  &38.4	&50.4	&13.4	&31.7	&52.1	&74.2	&82.0	&84.0	&35.8	&36.6	&44.3	&55.4	&53.7	&46.6

 \\
                && HDC${}^\dag$\cite{wei2024semi} &41.1	&54.2	&18.5	&43.6	&55.0	&77.9	&81.0	&83.2	&40.2	&41.4	&\textcolor{blue}{\textbf{47.2}}	&\textcolor{blue}{\textbf{60.1}}	&57.0	&50.3

\\
				% \cline{3-17}
                \hhline{~———----------------}
                &\multirow{2.3}[0]{*}{\rotatebox{90}{Ours}} 
				& \textbf{SPRED}  &48.8	&60.4	&10.6	&26.4	&50.5	&72.3	&82.4	&84.1	&41.9	&42.6	&46.8	&57.2	&55.2	&47.9
 \\
                &&\textbf{SPRED${}^\ddag$} &47.2	&60.0	&16.8	&40.9	&57.8	&78.4	&80.4	&82.2	&48.2	&49.9	&\textcolor{red}{\textbf{50.1}}	&\textcolor{red}{\textbf{62.3}}	&\textcolor{red}{\textbf{61.1}}	&\textcolor{red}{\textbf{54.1}}

\\

				\hline
				\multirow{11}[1]{*}{\rotatebox{90}{20\%}} &\multirow{3.3}[0]{*}{\rotatebox{90}{LReID}} &
                PatchKD~\cite{sun2022patch} &28.2	&37.4	&2.6	&7.3	&20.1	&40.7	&66.3	&70.5	&5.8	&4.7	&24.6	&32.1	&27.9	&23.1

 \\		
                % & & CKP~\cite{xu2024lstkc} &48.1  &72.7  &78.8  &81.4  &39.2  &53.4  &16.2  &38.5  &29.7  &29.4  &42.4  &55.1  &52.2  &43.8  \\
				& & DKP~\cite{xu2024distribution}  &21.7	&31.5	&7.2	&21.3	&28.0	&51.6	&73.8	&76.7	&6.7	&5.5	&27.5	&37.3	&40.0	&33.3

  \\

				& & LSTKC~\cite{xu2024lstkc} &24.3	&33.8	&9.3	&23.8	&39.5	&62.6	&76.7	&79.6	&18.7	&17.3	&33.7	&43.4	&44.3	&37.2

  \\

				% \cline{2-17}
                \hhline{~----------------}
				&\multirow{5.3}[0]{*}{\rotatebox{90}{SSL+LReID}} & 
				CDMAD${}^\dag$~\cite{lee2024cdmad} &20.7	&32.7	&5.8	&19.5	&26.3	&53.6	&73.6	&77.6	&9.1	&7.4	&27.1	&38.2	&43.5	&37.8

  \\
				&& ShrinkM${}^\dag$\cite{yang2023shrinking}  &31.9	&45.8	&15.1	&38.9	&44.6	&70.4	&73.6	&76.6	&23.3	&22.9	&37.7	&50.9	&51.2	&44.5

 \\
                &&SimMV2${}^\dag$~\cite{zheng2023simmatchv2}   &26.6	&38.8	&10.4	&28.3	&39.3	&62.5	&67.5	&69.9	&19.1	&18.4	&32.6	&43.6	&46.4	&38.7

  \\
                && DPIS${}^\dag$\cite{shi2023dual} &25.1	&35.9	&9.7	&25.0	&41.0	&64.1	&78.7	&81.5	&19.2	&17.3	&34.7	&44.8	&46.1	&39.1

 \\
                && HDC${}^\dag$\cite{wei2024semi} &28.6	&41.1	&12.4	&32.6	&43.5	&66.9	&76.0	&78.5	&24.3	&23.9	&37.0	&48.6	&50.1	&42.3

 \\
				% \cline{3-17}
                \hhline{~———----------------}
                &\multirow{2.3}[0]{*}{\rotatebox{90}{Ours}}
				& \textbf{SPRED}  &44.8 &56.3 &9.6 &24.1 &47.0 &69.0 &82.1 &84.1 &31.1 &32.2 &\textcolor{blue}{\textbf{42.9}} &\textcolor{blue}{\textbf{53.1}} &\textcolor{blue}{\textbf{52.3}} &\textcolor{blue}{\textbf{44.7}}
   \\
                &&\textbf{SPRED${}^\ddag$} &42.2 &55.5 &16.2 &40.7 &54.3 &76.7 &78.7 &80.6 &40.3 &40.5 &\textcolor{red}{\textbf{46.3}} &\textcolor{red}{\textbf{58.8}} &\textcolor{red}{\textbf{57.7}} &\textcolor{red}{\textbf{50.5}}
  \\
                \hline
				\multirow{11}[1]{*}{\rotatebox{90}{10\%}} &\multirow{3.3}[0]{*}{\rotatebox{90}{LReID}} &
                PatchKD~\cite{sun2022patch} &15.6 &22.8 &1.4 &4.2 &12.0 &28.9 &61.8 &65.4 &4.0 &3.2 &19.0 &24.9 &22.5 &18.1

 \\		
                % & & CKP~\cite{xu2024lstkc}  &40.4  &67.0  &75.3  &78.3  &33.2  &46.2  &13.5  &34.5  &24.3  &24.2  &37.3  &50.0  &47.9  &40.8  \\
				& & DKP~\cite{xu2024distribution}   &12.5 &19.4 &5.1 &15.4 &18.7 &40.2 &69.2 &72.3 &5.0 &3.8 &22.1 &30.2 &34.6 &28.5
   \\
				& & LSTKC~\cite{xu2024lstkc}  &17.4 &24.9 &6.3 &17.1 &27.3 &49.6 &73.1 &76.0 &13.6 &11.1 &27.5 &35.7 &38.9 &32.0
 \\
				% \cline{2-17}
                \hhline{~----------------}
				&\multirow{5.3}[0]{*}{\rotatebox{90}{SSL+LReID}} & 
				CDMAD${}^\dag$~\cite{lee2024cdmad} &17.8 &28.9 &6.0 &20.2 &25.9 &52.6 &72.0 &75.6 &9.3 &7.8 &26.2 &37.0 &41.7 &35.0
 \\
				&& ShrinkM${}^\dag$\cite{yang2023shrinking}   &24.8 &37.8 &11.7 &32.2 &33.9 &59.6 &71.0 &73.9 &19.1 &18.6 &32.1 &44.4 &46.9 &39.6
  \\
                &&SimMV2${}^\dag$~\cite{zheng2023simmatchv2}  &22.4 &33.2 &8.3 &23.6 &31.0 &53.9 &65.7 &69.2 &14.7 &14.3 &28.4 &38.8 &42.1 &34.7

   \\
                && DPIS${}^\dag$\cite{shi2023dual} &17.7 &26.1 &6.2 &17.0 &28.5 &51.1 &74.1 &77.2 &14.1 &12.4 &28.1 &36.8 &40.7 &33.4
 \\
                && HDC${}^\dag$\cite{wei2024semi} &17.5 &25.7 &5.7 &17.8 &26.0 &48.8 &71.5 &75.0 &13.4 &11.7 &26.8 &35.8 &41.3 &34.1
 \\
				% \cline{3-17}	
                \hhline{~———----------------}
                &\multirow{2.3}[0]{*}{\rotatebox{90}{Ours}}
                  & \textbf{SPRED}  &36.3 &47.3 &9.9 &24.9 &47.4 &68.6 &81.3 &83.2 &34.8 &34.9 &\textcolor{blue}{\textbf{41.9}} &\textcolor{blue}{\textbf{51.8}} &\textcolor{blue}{\textbf{50.7}} &\textcolor{blue}{\textbf{43.4}}
 \\
                 & & \textbf{SPRED${}^\ddag$}  &40.6 &54.1 &16.3 &40.3 &52.9 &76.1 &78.5 &81.2 &38.6 &39.2 &\textcolor{red}{\textbf{45.4}} &\textcolor{red}{\textbf{58.2}} &\textcolor{red}{\textbf{56.7}} &\textcolor{red}{\textbf{49.0}}
  \\
				\hline
        \end{tabular}%
    }
    \caption{Results comparison under different label rate $r$ under Training Order-2: LPW$\rightarrow$MSMT17$\rightarrow$Market-1501$\rightarrow$CUHK-SYSU$\rightarrow$CUHK03. ${}^\dag$ indicates the state-of-the-art SSL method is integrated with the anti-forgetting designs of LSTKC. ${}^\ddag$ represents adopting the data augmentation strategy of ShrinkM.}	        
    \label{tab:setting2}%
\end{table*}%

\section{Experimental Results on Different Dataset Orders}
In our main paper, the default training dataset order follows Market-1501$\rightarrow$CUHK-SYSU$\rightarrow$LPW$\rightarrow$MSMT17 $\rightarrow$CUHK03, denoted as Training Training Order-1~\footnote{(Training Order-1) Market-1501$\rightarrow$CUHK-SYSU$\rightarrow$LPW $\rightarrow$ MSMT17$\rightarrow$CUHK03}. In this section, we extend the evaluation by comparing our method with existing approaches under another training order: LPW$\rightarrow$MSMT17$\rightarrow$Market-1501$\rightarrow$CUHK-SYSU $\rightarrow$CUHK03, referred to as Training Order-2~\footnote{(Training Order-2) LPW$\rightarrow$MSMT17$\rightarrow$Market-1501$\rightarrow$ CUHK-SYSU 
$\rightarrow$CUHK03}. The results are summarized in \cref{tab:setting2}.

\textbf{Compared with LReID Methods}: 
Under Training Order-2, our proposed SPRED consistently surpasses existing LReID methods in both Seen-Avg and UnSeen-Avg mAP/R@1 across different label rates. Specifically, compared to the state-of-the-art LSTKC, SPRED achieves increasing improvements of \textbf{2.6\%/2.2\%}, \textbf{9.2\%/9.7\%}, and \textbf{14.4\%/16.1\%} in Seen-Avg mAP/R@1 as the label rate drops from 50\% to 10\%. Besides, SPRED obtains \textbf{1.9\%/1.9\%}, \textbf{8.0\%/7.5\%}, and \textbf{11.8\%/11.4\%} improvements in UnSeen-Avg mAP/R@1. These results highlight SPRED's superior adaptability to scenarios with reduced labeling rates, outperforming current state-of-the-art LReID methods in low-labeling regimes.
\begin{table*}[htbp]
   \centering
  \setlength{\tabcolsep}{0.7mm}{
    \begin{tabular}{llccccccccccc>{\columncolor{seen_back}}c>{\columncolor{seen_back}}c>{\columncolor{unseen_back}}c>{\columncolor{unseen_back}}cccc}   
    \hline
   \multirow{2}[1]{*}{$r$}&\multirow{2}[1]{*}{Method}&\multirow{2}[1]{*}{Publication} & \multicolumn{2}{c}{Market-1501} & \multicolumn{2}{c}{CUHK-SYSU} & \multicolumn{2}{c}{LPW} & \multicolumn{2}{c}{MSMT17} & \multicolumn{2}{c}{CUHK03} & \multicolumn{2}{>{\columncolor{seen_back}}c}{\textbf{Seen-Avg}} & \multicolumn{2}{>{\columncolor{unseen_back}}c}{\textbf{UnSeen-Avg}} \\
\hhline{~~~--------------}
&&& mAP   & R@1   & mAP   & R@1   & mAP   & R@1   & mAP   & R@1   & mAP   & R@1   & mAP   & R@1   & mAP   & R@1 \\
\hline
\multirow{5}[0]{*}{\rotatebox{90}{100\%} } 
%&CRL &WACV'21          & 58.0  & 78.2  & 72.5  & 75.1  & 28.3  & 45.2  & 6.0   & 15.8  & 37.4  & 39.8  & 40.5  & 50.8  & 47.8  & 43.5  \\
    %&AKA &CVPR 2021  & 51.2  & 72.0  & 47.5  & 45.1  & 18.7  & 33.1  & 16.4  & 37.6  & 27.7  & 27.6  & 32.3  & 43.1  & 44.3  & 40.4  \\
&
PatchKD &  MM'22       &71.6	&87.7	&77.0	&79.6	&33.2	&41.9	&7.0	&18.5	&29.5	&30.4	&43.7	&51.6	&47.8	&41.4
     \\    
%&MEGE&T-PAMI'23  & 39.0    & 61.6  & 73.3  & 76.6  & 16.9  & 30.3  & 4.6   & 13.4  & 36.4  & 37.1  & 34.0    & 43.8  & 47.7  & 44.0 \\
% &ConRFL &PR 2023      & 59.2  & 78.3  &\textcolor{blue}{\textbf{82.1}}  &\textcolor{blue}{\textbf{84.3}}  & 45.6  & 61.8  & 12.6  & 30.4  &\textcolor{blue}{\textbf{51.7}}  &\textcolor{blue}{\textbf{53.8}} & 50.2  & 61.7  & 57.4  &\textcolor{blue}{\textbf{52.3}} \\
% &CKP& ACM MM'24 &59.1 &	81.3 	&83.2 &	85.1 &	48.2 &	59.6 	&18.0 &	40.6 	&41.9 &	43.4 	&50.1 &	62.0 &58.2 &	51.2  \\
&LSTKC &AAAI'24   &57.0 	&78.6 	&82.9 	&84.9 	&47.2 	&58.4 	&18.4 	&41.1 	&42.3 	&43.7 	&49.6	&61.3	&57.8 	&50.2
  \\
&DKP &CVPR'24     &60.0 	&80.3 	&84.1 	&85.9 	&46.0 	&57.9 	&17.7 	&38.5 	&41.0 	&41.4 	&49.8	&60.8	&57.5 	&50.7
    \\	
\hhline{~----------------}
% \cline{2-17}
&\textbf{SPRED} & This Paper  &63.1 	&81.7 	&83.2 	&84.8 	&50.6 	&60.7 	&15.2 	&34.5 	&48.6 	&50.0 	&\textcolor{blue}{\textbf{52.1}}	&\textcolor{blue}{\textbf{62.3	}}&\textcolor{blue}{\textbf{58.7 }}	&\textcolor{blue}{\textbf{50.7}}
\\
&\textbf{SPRED${}^{\ddag}$} & This Paper   &65.0 	&83.3 	&81.8 	&83.6 	&51.1 	&63.1 	&21.4 	&47.5 	&52.6 	&53.6 	&\textcolor{red}{\textbf{54.4}}	&\textcolor{red}{\textbf{66.2}}	&\textcolor{red}{\textbf{62.8}} 	&\textcolor{red}{\textbf{55.8}}
 \\
    \hline  
    \hline
    \multirow{2}[1]{*}{$r$}&\multirow{2}[1]{*}{Method}&\multirow{2}[1]{*}{Publication} &  \multicolumn{2}{c}{LPW} & \multicolumn{2}{c}{MSMT17} & \multicolumn{2}{c}{Market-1501} & \multicolumn{2}{c}{CUHK-SYSU} & \multicolumn{2}{c}{CUHK03} & \multicolumn{2}{>{\columncolor{seen_back}}c}{\textbf{Seen-Avg}} & \multicolumn{2}{>{\columncolor{unseen_back}}c}{\textbf{UnSeen-Avg}} \\
\hhline{~~~--------------}
&&& mAP   & R@1   & mAP   & R@1   & mAP   & R@1   & mAP   & R@1   & mAP   & R@1   & mAP   & R@1   & mAP   & R@1 \\
\hline
\multirow{5}[0]{*}{\rotatebox{90}{100\%} } 
%&CRL &  WACV 2021    & 43.5  & 63.1  & 4.8   & 13.7  & 35.0  & 59.8  & 70.0  & 72.8  & 34.5  & 36.8  & 37.6  & 49.2  & 45.3  & 41.4  \\
   % &AKA & CVPR 2021  & 32.5  & 49.7  & -     & -     & -     & -     & -     & -     & -     & -     & -     & -     & 40.8  & 37.2  \\
    % &AKA$\dag$ & CVPR 2021   & 42.2  & 60.1  & 5.4   & 15.1  & 37.2  & 59.8  & 71.2  & 73.9  & 36.9  & 37.9  & 38.6  & 49.4  & 46.0  & 41.7  \\
    &PatchKD &  MM 2022      &58.0	&69.0	&6.3	&16.7	&46.3	&70.6	&75.7	&78.5	&29.6	&30.2	&43.2	&53.0	&45.3	&38.5
    \\  
   % &MEGE& T-PAMI 2023  & 21.6  & 35.5  & 3.0     & 9.3   & 25.0    & 49.8  & 69.9  & 73.1  & 34.7  & 35.1  & 30.8  & 40.6  & 44.3  & 41.1 \\ 
    % &ConRFL& PR 2023 &34.4  & 51.3  & 7.6   & 20.1  &\textcolor{blue}{\textbf{61.6}}  & 80.4  & 82.8  &\textcolor{blue}{\textbf{85.1}}  &\textcolor{blue}{\textbf{49.0}}    &\textcolor{blue}{\textbf{50.1}}  & 47.1  & 57.4  & 57.9  &\textcolor{blue}{\textbf{53.4}}\\       
    % &CKP&ACM MM 2024  &48.3 	&60.0 &	15.3 &	36.2 	&55.8 	&78.1 &	83.8 &	85.6 &	41.7 	&42.7 &	49.0 	&60.5 	&58.0 	&50.5  \\
    &LSTKC& AAAI 2024 &46.7 	&57.6 	&14.9 	&33.9 	&56.5 	&78.0 	&84.0 	&86.1 	&42.1 	&43.7 	&48.8	&59.9	&57.4 	&49.5
  \\    
    &DKP& CVPR 2024   &49.5 	&61.4 	&14.1 	&32.6 	&60.3 	&80.6 	&84.5 	&86.4 	&43.6 	&43.7 	&\textcolor{blue}{\textbf{50.4}}	&\textcolor{blue}{\textbf{60.9}}	&\textcolor{blue}{\textbf{59.5}} 	&\textcolor{blue}{\textbf{52.4}}
 \\	
\hhline{~----------------}
% \cline{2-17}
&\textbf{SPRED} & This Paper &51.8 	&62.4 	&10.4 	&26.2 	&56.8 	&77.0 	&84.5 	&85.9 	&42.9 	&43.8 	&49.3	&59.1	&57.1 	&49.1
\\
&\textbf{SPRED${}^{\ddag}$} & This Paper &51.6 	&63.4 	&16.4 	&40.1 	&61.9 	&82.5 	&82.7 	&84.4 	&49.4 	&51.0 	&\textcolor{red}{\textbf{52.4}}	&\textcolor{red}{\textbf{64.3}}	&\textcolor{red}{\textbf{60.8}} 	&\textcolor{red}{\textbf{53.6}}
 \\
\hline
    \end{tabular}%
   }
 \raggedright
\caption{
Comparison with LReID methods under label rate $r$=100\% on Training Order-1: Market-1501$\rightarrow$CUHK-SYSU$\rightarrow$LPW$\rightarrow$ MSMT17$\rightarrow$CUHK03 and Training Order-2: LPW$\rightarrow$MSMT17$\rightarrow$Market-1501$\rightarrow$CUHK-SYSU$\rightarrow$CUHK03.
  }
\label{tab:label-rate-100}%
\end{table*}%

\begin{table*}[htbp]
   \centering
  \setlength{\tabcolsep}{0.95mm}{
    \begin{tabular}{llccccccccccc>{\columncolor{seen_back}}c>{\columncolor{seen_back}}c>{\columncolor{unseen_back}}c>{\columncolor{unseen_back}}cccc}   
    \hline
   \multirow{2}[1]{*}{$r$}&\multirow{2}[1]{*}{Method}&\multirow{2}[1]{*}{Publication} & \multicolumn{2}{c}{Market-1501} & \multicolumn{2}{c}{CUHK-SYSU} & \multicolumn{2}{c}{LPW} & \multicolumn{2}{c}{MSMT17} & \multicolumn{2}{c}{CUHK03} & \multicolumn{2}{>{\columncolor{seen_back}}c}{\textbf{Seen-Avg}} & \multicolumn{2}{>{\columncolor{unseen_back}}c}{\textbf{UnSeen-Avg}} \\
\hhline{~~~--------------}
&&& mAP   & R@1   & mAP   & R@1   & mAP   & R@1   & mAP   & R@1   & mAP   & R@1   & mAP   & R@1   & mAP   & R@1 \\
\hline
\multirow{4}[0]{*}{\rotatebox{90}{50\%} } 
&
CKP &  MM'24       &55.2  & 78.5  & 81.0  & 82.8  & 44.3  & 56.9  & 17.6  & 40.5  & 39.2  & 40.9  & 47.4  & 59.9  & 55.7  & 48.1  \\

&LDC &ECCV'24   &23.4  & 50.1  & 61.1  & 65.9  & 19.3  & 32.7  & 8.8   & 29.4  & 17.7  & 17.8  & 26.0  & 39.2  & 38.7  & 32.3  \\
\hhline{~----------------}
% \cline{2-17}
& \textbf{SPRED}&This Paper &59.8	&79.7	&82.8	&84.6	&49.7	&60.2	&17.4	&38.2	&40.6	&41.3	&\textcolor{blue}{\textbf{50.1}}	&\textcolor{blue}{\textbf{60.8}}	&\textcolor{blue}{\textbf{57.8}}	&\textcolor{blue}{\textbf{50.3}}
  \\
                &\textbf{SPRED${}^\ddag$} &This Paper &59.4	&80.7	&79.6	&81.5	&47.3	&60.2	&21.6	&47.3	&48.4	&49.4	&\textcolor{red}{\textbf{51.3}}	&\textcolor{red}{\textbf{63.8}}	&\textcolor{red}{\textbf{60.7}}	&\textcolor{red}{\textbf{53.9}}
 \\
    \hline  
\multirow{4}[0]{*}{\rotatebox{90}{20\%} } 
&
CKP &  MM'24       &48.1  & 72.7  & 78.8  & 81.4  & 39.2  & 53.4  & 16.2  & 38.5  & 29.7  & 29.4  & 42.4  & 55.1  & 52.2  & 43.8  \\

&LDC &ECCV'24   &19.1  & 44.8  & 56.0  & 60.8  & 15.0  & 26.6  & 6.5   & 23.1  & 8.2   & 6.8   & 21.0  & 32.4  & 32.3  & 26.5  \\

\hhline{~----------------}
% \cline{2-17}
& \textbf{SPRED}  &This Paper &54.7	&75.7	&81.1	&83.0	&45.0	&56.9	&16.1	&36.3	&35.8	&35.4	&\textcolor{blue}{\textbf{46.5}}	&\textcolor{blue}{\textbf{57.5}}	&\textcolor{blue}{\textbf{55.7}}	&\textcolor{blue}{\textbf{48.4}}

   \\
                &\textbf{SPRED${}^\ddag$} &This Paper &54.6	&77.1	&78.5	&80.5	&44.9	&58.9	&21.3	&47.5	&38.6	&38.2	&\textcolor{red}{\textbf{47.6}}	&\textcolor{red}{\textbf{60.4}}	&\textcolor{red}{\textbf{58.9}}	&\textcolor{red}{\textbf{51.5}}
\\
\hline
\multirow{4}[0]{*}{\rotatebox{90}{10\%} } 
&
CKP &  MM'24       &40.4  & 67.0  & 75.3  & 78.3  & 33.2  & 46.2  & 13.5  & 34.5  & 24.3  & 24.2  & 37.3  & 50.0  & 47.9  & 40.8  \\

&LDC &ECCV'24   &12.3  & 33.3  & 52.4  & 56.6  & 8.9   & 16.8  & 3.9   & 16.0  & 5.6   & 5.1   & 16.6  & 25.5  & 28.4  & 22.6  \\
\hhline{~----------------}
% \cline{2-17}

                  & \textbf{SPRED}  &This Paper &45.9	&69.4	&79.3	&81.6	&40.0	&51.1	&14.2	&33.0	&36.7	&37.1	&\textcolor{blue}{\textbf{43.2}}	&\textcolor{blue}{\textbf{54.4}}	&\textcolor{blue}{\textbf{51.7}}	&\textcolor{blue}{\textbf{44.6}}

 \\
                 & \textbf{SPRED${}^\ddag$}  &This Paper &49.5	&74.0	&76.5	&79.1	&40.9	&55.4	&19.4	&45.3	&37.5	&38.4	&\textcolor{red}{\textbf{44.8}}	&\textcolor{red}{\textbf{58.4}}	&\textcolor{red}{\textbf{55.9}}	&\textcolor{red}{\textbf{48.7}}\\
\hline
  
    \end{tabular}%
   }
 \raggedright
\caption{
Comparison with NoisyLReID and SSLL methods on Training Order-1: Market-1501$\rightarrow$CUHK-SYSU$\rightarrow$LPW$\rightarrow$ MSMT17$\rightarrow$CUHK03.
  }
\label{tab:noisy}%
\end{table*}%

\textbf{Compared with SSL+LReID Methods}: 
When adopting the traditional data augmentation configuration, our  SPRED model demonstrates significantly superior performance, particularly at a low label rate of $r$=10\%. 
When integrating the data augmentation strategy from SSL, the enhanced SPRED${}^\ddag$ model outperforms all existing methods in both Seen-Avg mAP/R@1 and UnSeen-Avg mAP/R@1 metrics across all label rates. Specifically, SPRED${}^\ddag$ achieves notable improvements in Seen-Avg mAP/R@1 of \textbf{2.9\%/2.2\%}, \textbf{9.3\%/10.2\%} and \textbf{13.3\%/13.8\%} at label rates of 50\%, 20\%, and 10\%, respectively. In addition, in UnSeen-Avg mAP/R@1, improvements of \textbf{3.4\%/3.2\%}, \textbf{6.5\%/6.0\%} and \textbf{9.8\%/9.4\%} are observed at label rates of 50\%, 20\%, and 10\%, separately.
These results underscore the effectiveness of our neighbor prototype labeling and dual-knowledge utilization mechanisms, which significantly enhance pseudo-label quality and improve the model's capability to acquire new knowledge.
\begin{figure*}[ht]
    \centering
	\includegraphics[width=0.9\linewidth]{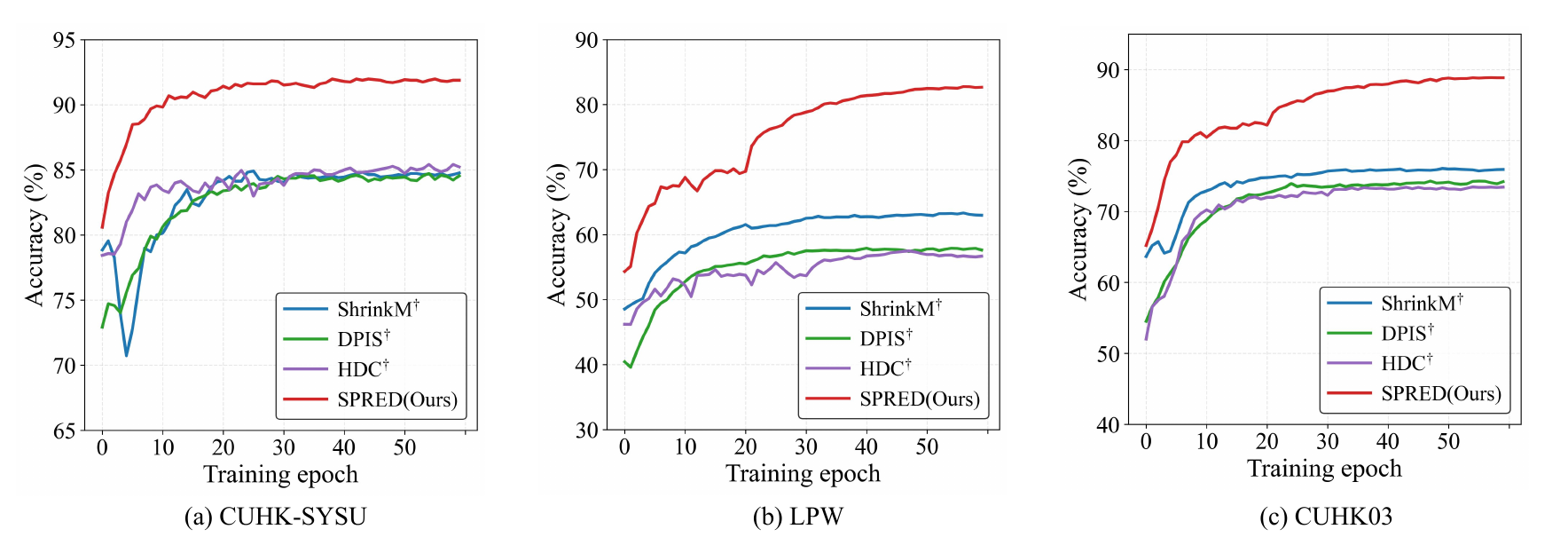}
 % \vspace{-18pt}
        \caption{\label{fig:label_acc_more}  Label prediction accuracy of different methods. The compared methods are trained on Training Order-1 with label rate $r$=10\%}
\end{figure*}

\begin{figure}[t]
    \centering
    % \vspace{3pt}
	\includegraphics[width=1\linewidth]{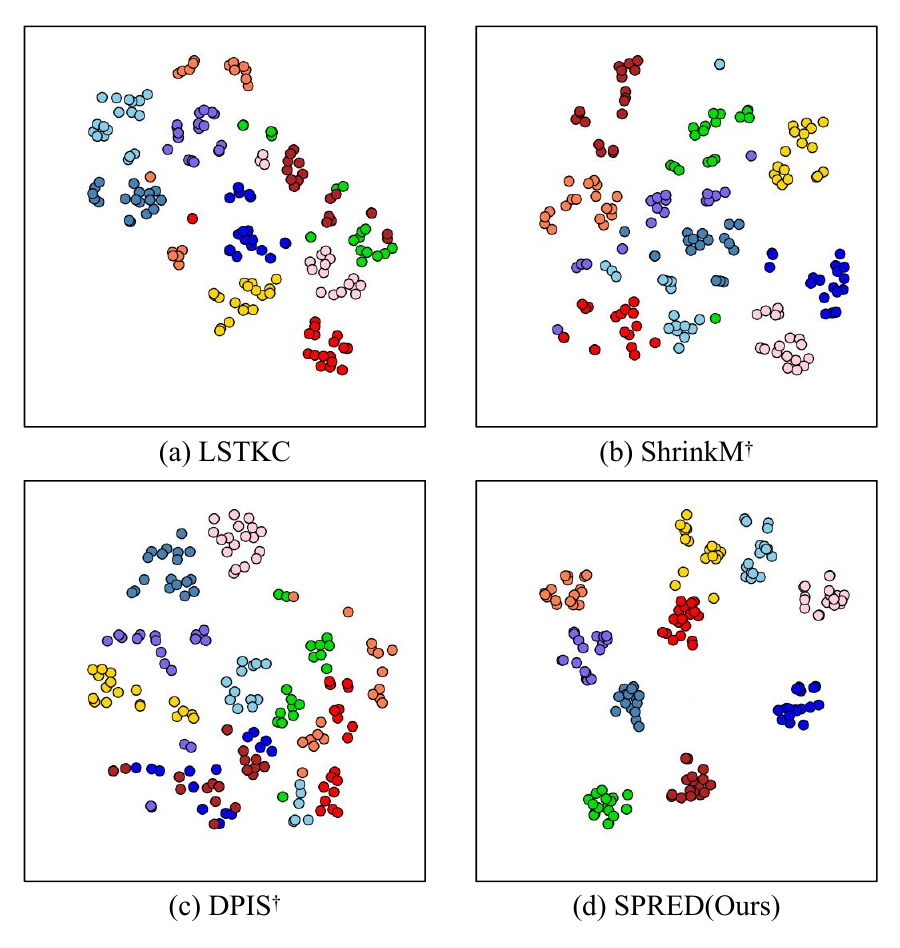}
 % \vspace{-18pt}
        \caption{\label{fig:test-feat} Visualization of test feature distribution. The compared methods are trained on Training Order-1 with label rate $r$=10\%. Each color represents an identity.}
        % \vspace{-18pt}
\end{figure}

\section{Experimental Results Under Label Rate $r$=100\%}
We also evaluate our method under the fully-labeled scenario, where the label rate $r$=100\%. As shown in \cref{tab:label-rate-100}, our SPRED achieves competitive results with the latest LReID methods across both training orders.
Additionally, the enhanced SPRED${}^\ddag$ surpasses all existing methods with significant margins. Specifically, SPRED${}^\ddag$ outperforms the state-of-the-art DKP by \textbf{4.6\%/5.4\%} and \textbf{2.0\%/3.4\%} in Seen-Avg mAP/R@1 across the two training orders. For UnSeen-Avg mAP/R@1, SPRED${}^\ddag$ achieves improvements of \textbf{4.1\%/5.1\%} and \textbf{1.3\%/1.2\%} over DKP. These results verify the adaptability of our method to varying scenarios
% These results highlight the adaptability of our method to the fully supervised scenarios.

Note that while DKP~\cite{xu2024distribution} outperforms our SPRED at the label rate of 100\% under training order-2 due to its distribution modeling design, this approach relies heavily on abundant labels to ensure reliable identity distribution modeling. In scenarios with insufficient labeled samples, the learned distributions become less reliable, leading to degraded performance, as demonstrated in \cref{tab:setting2}. Consequently, our method is more suitable in situations where the labeling source is restricted.

\section{Quantitative comparisons with state-of-the-art NoisyLReID and SSLL methods}
We also compare our proposed method with the latest Noisy Lifelong Person Re-Identification (NoisyLReID) method CKP~\cite{xu2024mitigate} by assigning all the unlabeled data with the same label, serving as noisy data. Besides, the exemplar-free SSLL method LDC~\cite{gomez2024exemplar} is also compared (with sequential fine-tune as the baseline). The results in Tab.~\ref{tab:noisy} show that our SPRED and SPRED${}^\ddag$ outperform both competitors significantly across all label rates.

\section{Additional Visualization on Prediction and Features}

\textbf{More Identity Prediction Capacity Visualization}. 
Beyond the prediction accuracy curves presented in the main paper, we provide additional visualizations for CUHK-SYSU, LPW, and CUHK03 datasets in \cref{fig:label_acc_more}. The results show that SPRED consistently achieves higher prediction accuracy throughout training epochs. This can be attributed to the cyclically evolved pseudo-labeling mechanism, where pseudo-label prediction and purification iteratively reinforce each other.

\textbf{Testing Feature Visualization}. 
We visualize the test data features in \cref{fig:test-feat}, where each point represents the feature of a test sample, with each color representing an identity. The results illustrate that features extracted by SPRED exhibit better clustering compared to existing methods. This improvement arises from the dual-knowledge-guided pseudo-label purification mechanism, which enhances pseudo-label quality and facilitates effective knowledge learning.

% \section{Retrieval Results Visualization}
% In this section, we compare the person retrieval performance of our model and the state-of-the-art ShrinkM${}^\dag$ under the Semi-LReID benchmark with a label rate of $r$=10\%. Visualizations for seen datasets Market-1501 (\cref{fig:market}) and CUHK-SYSU (\cref{fig:cuhk-sysu}) highlight the acquisition and retention of knowledge, while unseen datasets CUHK01 (\cref{fig:cuhk01}) and VIPeR (\cref{fig:viper}) demonstrate generalization capability.

% In the visualizations, black-boxed images represent input query images, while green- and red-boxed images indicate correct and incorrect retrievals, respectively, arranged in descending order of matching scores. The results show that SPRED consistently retrieves more accurate matches, even under challenging conditions such as variations in pose, viewpoint, and scene. This demonstrates the robust knowledge acquisition and accumulation capabilities of our proposed method.

% {
%     \small
%     \bibliographystyle{ieeenat_fullname}
%     \bibliography{main}
% }

\end{document}